\documentclass{article} 
\usepackage{iclr2027_conference,times}

\usepackage{amsmath,amsfonts,bm}

\def\eqref#1{equation~\ref{#1}}

\def\1{\bm{1}}

\DeclareMathAlphabet{\mathsfit}{\encodingdefault}{\sfdefault}{m}{sl}
\SetMathAlphabet{\mathsfit}{bold}{\encodingdefault}{\sfdefault}{bx}{n}

\usepackage[hidelinks]{hyperref}
\usepackage{comment}
\usepackage{epsfig}
\usepackage{graphicx}
\usepackage{array}
\usepackage{xspace}
\usepackage{microtype}
\usepackage{url}
\usepackage{amsmath,amsfonts,amssymb,bm,amsthm}
\usepackage{graphicx}
\usepackage{booktabs}
\usepackage{array}
\usepackage{url}
\usepackage{wrapfig}
\usepackage{multirow,multicol}
\usepackage{enumitem}
\usepackage{xcolor}
\usepackage{xspace} 
\usepackage{enumitem}
\usepackage{times}
\usepackage{subcaption}
\usepackage{cleveref}

\definecolor{wacvblue}{rgb}{0.21,0.49,0.74}

\newcommand{\DEC}{\textsc{Dec}\xspace}
\newcommand{\PROB}{\textsc{Prob}\xspace}

\newcolumntype{Y}{>{\centering\arraybackslash}X}
\setlist[itemize]{leftmargin=*,topsep=2pt,itemsep=1pt,parsep=0pt}

\title{A VLM Answer Is Not an Anomaly Score: \\
Rank Compression Across Image and \\
Video Anomaly Detection}

\author{
Inpyo Song \\
SungKyunKwan University \\
South Korea \\
\texttt{songinpyo@skku.edu}
\And
Jangwon Lee \\
SungKyunKwan University \\
South Korea \\
\texttt{leejang@skku.edu}
}

\newif\ifarxiv
\arxivtrue          

\ifarxiv
  \iclrfinalcopy    
\fi

\begin{document}

\maketitle

\ifarxiv
  \fancyhead{}      
\fi

\maketitle

\begin{abstract}
Anomaly detection aims to identify observations that deviate from normal patterns.
Recent work has increasingly used pretrained vision--language models (VLMs) for training-free image and video anomaly detection without task-specific retraining.
Anomaly detection is commonly evaluated by how well anomaly scores rank anomalous images or video frames above normal ones.
Generative VLMs, however, assign probabilities to possible answers and then decode a single answer.
This decoding step may discard ordering information. We call this loss \textbf{decoded-answer rank compression}.
To isolate this effect, 
we compare two ways of scoring the same VLM output: one uses only the decoded answer, while the other computes a probability-weighted score over all possible answers.
Probability-weighted scoring consistently outperforms decoded-answer scoring 
across image and video anomaly detection benchmarks, 
using different VLMs and answer scales.
The mean gains range from 7.66 to 19.95 points on the primary metrics.
Most of this gap comes from decoded-answer ties.
Breaking decoded-answer ties with answer probabilities recovers at least 95\% of the average performance gap on every benchmark.
We further investigate how these ties affect evaluation.
We find that evaluating tied scores one input at a time makes the reported AUROC depend on input order,
and that linear PR interpolation can inflate reported performance.
These results reveal that both how VLM outputs are scored and how discrete anomaly scores are evaluated affect reported anomaly-detection performance.
\end{abstract}

\section{Introduction}
\label{sec:intro}
In computer vision, anomaly detection (AD) aims to identify deviations from normal patterns, such as defects in images or unusual events in videos.
Most image and video anomaly detectors learn a task-specific scoring function from normal examples or labeled anomalies~\citep{luo2025INP-Former,song2026bounding}.
Adapting these detectors to new environments or anomaly definitions typically requires additional training.
To avoid task-specific retraining, recent work uses pretrained vision--language models (VLMs) for training-free AD~\citep{xu2025anomalyov,zhang2025logsad,lavad}.
These methods are attractive because they can make anomaly judgments and 
provide natural language explanations without updating model parameters~\citep{anomalyruler,song2026instance}.

Image and video AD are commonly evaluated by how well anomaly scores rank anomalous images or video frames above normal ones, 
using metrics such as the area under the receiver operating characteristic curve (AUROC) and average precision (AP).
Generative VLMs, however, assign probabilities to possible answers and then decode a single answer, 
such as a Yes/No judgment or a numerical anomaly rating~\citep{zhang2024gpt,hofer2025kaputt,vera,anyanomaly}.
But if only the decoded answer is used as the anomaly score, inputs with different answer probabilities can receive the same score, discarding their relative ordering.

Consider the simplest case with two possible answers, Yes and No.
Suppose that ``Yes'' indicates an anomaly and that the VLM assigns ``Yes'' probabilities of 0.56 and 0.94 to two inputs.
Both inputs decode to ``Yes'', even though the answer probabilities rank the second input above the first.
We call this many-to-one loss of ordering \textbf{decoded-answer rank compression}.
We ask \emph{how much useful anomaly-ranking information is lost by decoded-answer scoring and where this loss comes from}.

\paragraph{Contributions.}
To answer these questions, we compare two scoring rules applied to the same VLM output.
Decoded-answer scoring assigns a score using only the decoded answer.
Probability-weighted scoring instead averages the answer values using their probabilities.
We keep the VLM, visual input, prompt, and answer probabilities fixed,
changing only how the same output distribution is converted into an anomaly score.

We evaluate decoded-answer and probability-weighted scoring across image and video AD benchmarks,
using different VLMs and answer scales.
Probability-weighted scoring consistently outperforms decoded-answer
scoring, with mean gains across VLMs and answer scales of 19.95
image-level AUROC points on VisA and 10.97 frame-level AP points
on the video benchmark XD-Violence.
The scoring-rule gap persists with larger VLMs, alternative prompts,
reversed answer polarity, and sampled decoding.

We find most of the gap comes from ties created by decoding.
To test this, 
we preserve the order between different decoded answers
and use answer probabilities only to break ties.
Across all seven answer scales, this tie-breaking recovers at least 95\%
of the average primary-metric gap on every benchmark.
Finer answer scales reduce rank compression, but
even on the finest 91-answer scale (0.0--9.0), 
the VLMs use only a small fraction of the available score values, 
leaving many inputs tied.

We also find that common evaluation choices can make this ranking
loss appear smaller.
Evaluating tied scores one at a time can make the reported ROC area
depend on input order~\citep{sultani2018}.
Trapezoidal PR-AUC can also inflate reported performance
by crediting precision values above those attainable by thresholding the anomaly scores~\citep{wu2020not,zou2022spotdiff}.
We therefore evaluate tied scores together and distinguish
trapezoidal PR-AUC from non-interpolated AP, which averages precision
attained at the score's thresholds~\citep{fawcett2006introduction,davis2006relationship}.

These results have two implications for VLM anomaly detection.
First, the scoring rule is part of detector design:
when answer probabilities are available, 
probability-weighted scoring retains useful ordering and improves anomaly ranking 
without changing the VLM or prompt.
Second, how these scores are evaluated also matters:
tie handling and PR-area calculation can materially affect reported anomaly-detection performance.


\section{Related Work}
\label{sec:related}

\paragraph{Vision--language models for visual anomaly detection.}
Recent work uses pretrained vision--language models for visual anomaly detection in 
both images and videos, 
reducing the need for task-specific training.
CLIP-based methods typically derive anomaly scores 
from the similarity 
between visual features and text descriptions 
of normality and abnormality~\citep{jeong2023winclip,zhou2024anomalyclip,li2024promptad,wu2024vadclip}.
More recent methods use generative VLMs to make anomaly judgments or provide explanations~\citep{gu2024anomalygpt,xu2025anomalyov,zhang2025logsad}.
Several image and video anomaly-detection methods convert decoded binary judgments or numerical ratings into anomaly scores~\citep{zhang2024gpt,hofer2025kaputt,anomalyruler,lavad,vera,anyanomaly,shao2025eventvad,li2026vadtree,song2026rethinking}.
Our work studies this scoring step directly: by holding the VLM answer distribution fixed, 
we measure how much anomaly-ranking information is lost when that distribution is reduced to a decoded answer.

\paragraph{Probability-based scoring with language models.}
Language-model probabilities have been used to construct scalar scores for document ranking, text evaluation, and answer verification~\citep{nogueira2020document,zhong2022unieval,liu2023g,zhuang2024beyond,zhang2025generative}.
Our question is different: how does converting a VLM answer distribution into an anomaly score affect anomaly ranking?
We use probability-weighted scoring as a comparison to measure this effect, 
analyze where the ranking differences arise, 
and study how they affect anomaly-detection evaluation.

\section{From a VLM Answer Distribution to an Anomaly Score}
\label{sec:method}
A VLM produces a distribution over possible answers, 
while anomaly detection requires a scalar anomaly score.
We compare two ways of obtaining this score from the same answer distribution.

\paragraph{Answer distribution.}
Let $x$ denote the visual input and let the prompt $q$ specify a finite ordered set of possible answers
$\mathcal{A}={a_0,\ldots,a_K}$.
We assign each answer $a$ an anomaly value $v(a)\in[0,1]$, with larger values indicating stronger anomaly judgments.
For Yes/No, for example, $v(\text{No})=0$ and $v(\text{Yes})=1$.

For each answer, we compute the log-probability of the complete answer string:
\begin{equation}
\ell_x(a)=\log p_\theta(a\mid x,q).
\end{equation}
For multi-token answers, $\ell_x(a)$ is the sum of the token log-probabilities.
We normalize these values over the possible answers:
\begin{equation}
\widetilde p_x(a)
=
\frac{\exp \ell_x(a)}
{\sum_{a'\in\mathcal{A}}\exp \ell_x(a')}.
\label{eq:normalizedprob}
\end{equation}
We refer to $\widetilde p_x$ as the answer distribution.

\paragraph{Two scoring rules.}
Decoded-answer scoring (\DEC) assigns the anomaly value of the most probable answer:
\begin{equation}
    a_x^*
    =
    \arg\max_{a\in\mathcal{A}}\ell_x(a),
    \qquad
    s_x^{\DEC}
    =
    v(a_x^*).
    \label{eq:dec}
\end{equation}
Probability-weighted scoring (\PROB) instead computes the expected anomaly value under the answer distribution:
\begin{equation}
    s_x^{\PROB}
    =
    \sum_{a\in\mathcal{A}}
    v(a)\widetilde p_x(a).
    \label{eq:prob}
\end{equation}
Thus the two scoring rules start from the same answer distribution and differ only in how to convert to anomaly score.

\paragraph{Decoded-answer rank compression.}
\DEC assigns the same anomaly score to all inputs with the same decoded answer.
With $|\mathcal{A}|$ possible answers, 
\DEC can therefore produce at most $|\mathcal{A}|$ distinct scores, 
even when the answer distributions differ across inputs.
We call this many-to-one loss of ordering \emph{decoded-answer rank compression}.

\section{Evaluation Setup}
\label{sec:setup}
To isolate the effect of the scoring rule, 
we compute \DEC and \PROB from the same answer distribution for every input.
The VLM, visual input, prompt, and inference precision are identical within each comparison.

\paragraph{Evaluation metrics.}
We evaluate anomaly scores using AUROC and non-interpolated average precision (AP).
For scores $s^+$ and $s^-$ of randomly sampled anomalous and normal examples, respectively,
\begin{equation}
    \operatorname{AUROC}(s)
    =
    \Pr(s^+ > s^-)
    +
    \frac{1}{2}\Pr(s^+ = s^-).
    \label{eq:auc}
\end{equation}
Thus, a tied anomalous--normal pair receives half credit~\citep{fawcett2006introduction}.

For AP, let $u_1>\cdots>u_J$ denote the distinct score values.
Let $p_j$ and $n_j$ be the numbers of anomalous and normal examples with score $u_j$, and define
\begin{equation}
    \mathrm{TP}_j=\sum_{i\le j}p_i,
    \qquad
    \mathrm{FP}_j=\sum_{i\le j}n_i,
    \qquad
    P=\sum_j p_j.
\end{equation}
We compute
\begin{equation}
    \operatorname{AP}(s)
    =
    \sum_{j=1}^{J}
    \frac{p_j}{P}
    \frac{\mathrm{TP}_j}
         {\mathrm{TP}_j+\mathrm{FP}_j}.
    \label{eq:ap}
\end{equation}
All examples with the same score are therefore evaluated together.
We distinguish this non-interpolated AP from trapezoidal PR-AUC, which linearly interpolates between precision--recall points~\citep{davis2006relationship,chen2024auprc}.

\paragraph{Benchmarks and evaluation protocol.}
We evaluate image anomaly detection on MVTec AD~\citep{bergmann2019mvtecad} and VisA~\citep{zou2022spotdiff}.
Each image is scored independently 
with its category name included in the prompt, 
and we report category-macro image-level AUROC (I-AUROC) and AP (I-AP).

For video anomaly detection, we evaluate UCF-Crime~\citep{sultani2018} and XD-Violence~\citep{wu2020not}.
Following VERA~\citep{vera}, we score non-overlapping 16-frame segments using frames sampled from their temporal context, 
assign each segment score to its frames, and compute frame-level AUROC and AP over the test set.
Additional details are given in \Cref{sec:supp_protocol}.

\begin{wraptable}{r}{0.37\columnwidth}
\centering
\vspace{-1.4em}
\caption{
Answer scales used in our experiments.
}
\label{tab:answer_scales}
\small
\begin{tabular}{lcc}
\toprule
Scale & Format & $|\mathcal{A}|$ \\
\midrule
Yes/No & binary      & 2  \\
0--1   & integer     & 2  \\
0--5   & integer     & 6  \\
0--9   & integer     & 10 \\
0--1   & one decimal & 11 \\
0--5   & one decimal & 51 \\
0--9   & one decimal & 91 \\
\bottomrule
\end{tabular}
\vspace{-1.5em}
\end{wraptable}

\paragraph{Models and answer scales.}
We evaluate four frozen VLMs: 
Qwen3-VL-8B~\citep{bai2025qwen3}, Qwen2.5-VL-7B~\citep{bai2025qwen25}, InternVL3.5-8B~\citep{zhu2025internvl3}, and MiniCPM-V-4.5-8B~\citep{yu2026minicpm}.
For each model, we use seven answer scales ranging from binary Yes/No to a 91-answer decimal scale as shown in~\Cref{tab:answer_scales}.
These scales let us compare \DEC and \PROB from coarse binary outputs to fine-grained numerical ratings.
Within each domain, the prompts describe anomalies in the same terms while changing the allowed answers and required response format.
The exact prompts are given in \Cref{sec:supp_prompts}.

\paragraph{Uncertainty estimates.}
We report paired 95\% nonparametric bootstrap confidence intervals using 10,000 replicates.
For image benchmarks, 
we resample test images within each category before macro-averaging, 
while for video benchmarks we resample complete videos.
Each bootstrap sample is shared between \DEC and \PROB, 
preserving the paired comparison between the two scoring rules.

\section{Evaluation}
\label{sec:evaluation}


\subsection{Main Results}
\label{sec:results}
\label{sec:raw}

\paragraph{Probability-weighted scoring improves performance across all tested settings.}
\PROB outperforms \DEC in all 112 model-scale-benchmark comparisons, 
covering every VLM and answer scale on all four benchmarks (\Cref{sec:supp_raw}).
\Cref{tab:raw_summary} summarizes the primary-metric results averaged over the four VLMs.
Across models and answer scales, the mean improvement ranges from 7.66 percentage points on UCF-Crime to 19.95 points on VisA.
The paired 95\% bootstrap interval for the mean improvement is strictly above zero on every benchmark .
Because each comparison is computed from the same answer distribution,
the observed difference comes from the scoring rule.

\begin{table*}[t]
\centering
\caption{
Primary-metric performance on 4 benchmarks averaged over four VLMs (percentage points).
Each \DEC and \PROB is computed from the same answer probability.
$\Delta$ is \PROB $-$ \DEC.
}
\label{tab:raw_summary}
\label{tab:image_raw_summary}
\small
\setlength{\tabcolsep}{3.0pt}
\resizebox{\textwidth}{!}{%
\begin{tabular}{lccrccrccrccr}
\toprule
& \multicolumn{3}{c}{MVTec AD, I-AUROC}
& \multicolumn{3}{c}{VisA, I-AUROC}
& \multicolumn{3}{c}{UCF-Crime, AUROC}
& \multicolumn{3}{c}{XD-Violence, AP} \\
\cmidrule(lr){2-4}\cmidrule(lr){5-7}\cmidrule(lr){8-10}\cmidrule(lr){11-13}
Scale ($|\mathcal{A}_k|$)
& \DEC & \PROB & $\Delta$
& \DEC & \PROB & $\Delta$
& \DEC & \PROB & $\Delta$
& \DEC & \PROB & $\Delta$ \\
\midrule
Yes/No (2)       & 69.63 & \textbf{88.94} & +19.31 & 58.44 & \textbf{81.06} & +22.62 & 72.20 & \textbf{85.32} & +13.12 & 49.44 & \textbf{68.98} & +19.54 \\
0--1 integer (2) & 61.04 & \textbf{89.67} & +28.63 & 54.21 & \textbf{82.89} & +28.68 & 73.69 & \textbf{85.42} & +11.73 & 50.53 & \textbf{68.08} & +17.55 \\
0--5 integer (6) & 77.07 & \textbf{89.81} & +12.74 & 66.54 & \textbf{83.97} & +17.43 & 79.05 & \textbf{84.76} &  +5.71 & 57.58 & \textbf{66.96} &  +9.38 \\
0--9 integer (10)& 77.19 & \textbf{89.48} & +12.28 & 67.10 & \textbf{83.79} & +16.70 & 78.71 & \textbf{84.38} &  +5.67 & 58.71 & \textbf{66.94} &  +8.23 \\
0--1 decimal (11)& 70.72 & \textbf{89.79} & +19.07 & 62.82 & \textbf{84.09} & +21.27 & 74.91 & \textbf{83.68} &  +8.78 & 57.32 & \textbf{65.67} &  +8.34 \\
0--5 decimal (51)& 78.30 & \textbf{89.74} & +11.44 & 66.33 & \textbf{83.37} & +17.04 & 80.37 & \textbf{84.52} &  +4.14 & 59.41 & \textbf{66.54} &  +7.12 \\
0--9 decimal (91)& 77.92 & \textbf{89.70} & +11.77 & 66.53 & \textbf{82.47} & +15.94 & 79.73 & \textbf{84.22} &  +4.49 & 60.22 & \textbf{66.85} &  +6.63 \\
\midrule
Mean over scales & 73.13 & \textbf{89.59} & +16.46 & 63.14 & \textbf{83.09} & +19.95 & 76.95 & \textbf{84.61} & +7.66 & 56.17 & \textbf{67.15} & +10.97 \\
\bottomrule
\end{tabular}%
}
\end{table*}

\paragraph{The improvement holds across image categories and video event classes.}
We next ask whether the benchmark-level gains are 
concentrated in only a few categories or event classes.
We evaluate this in the Yes/No setting, 
averaging each category- and class-level metric over the four VLMs.
\Cref{tab:image_per_category} shows that \PROB outperforms \DEC on both I-AUROC and I-AP in 
all 27 image categories across MVTec AD and VisA.
Similarly, \Cref{tab:video_per_class} shows higher AUROC and AP for \PROB in 
all 19 evaluated video event classes across UCF-Crime and XD-Violence.
The benchmark-level improvements are therefore not driven by 
a small subset of categories or event classes.

\paragraph{The improvement persists across model sizes.}
To test whether this result depends on model size, 
we repeat the Yes/No comparison with Qwen3-VL and InternVL3.5 models ranging from 2B to 38B parameters.
\Cref{tab:model_size} shows that \PROB outperforms \DEC on both metrics in every evaluated model and benchmark comparison.
On UCF-Crime, \PROB improves AUROC by 28.53 points for Qwen3-VL-2B and by 10.87 points for InternVL3.5-38B.
Thus, the improvement is not limited to the 7B and 8B models used in the main evaluation.

\paragraph{The improvement extends to low false-positive rates.}
AUROC and AP summarize performance across score thresholds, but do not characterize performance at any particular operating point.
We therefore compare attainable true-positive rate (TPR) at false-positive-rate (FPR) budgets of 1\%, 5\%, and 10\%.
\PROB achieves higher TPR on all four benchmarks at every tested FPR budget.
At the most restrictive 1\% budget, 
for example, the mean TPR increases from 34.91 to 61.36 on MVTec AD and from 1.00 to 16.46 on UCF-Crime.
Complete results are reported in \Cref{sec:supp_fixedfpr}.

\begin{figure*}[t]
\centering
\includegraphics[width=0.98\textwidth]{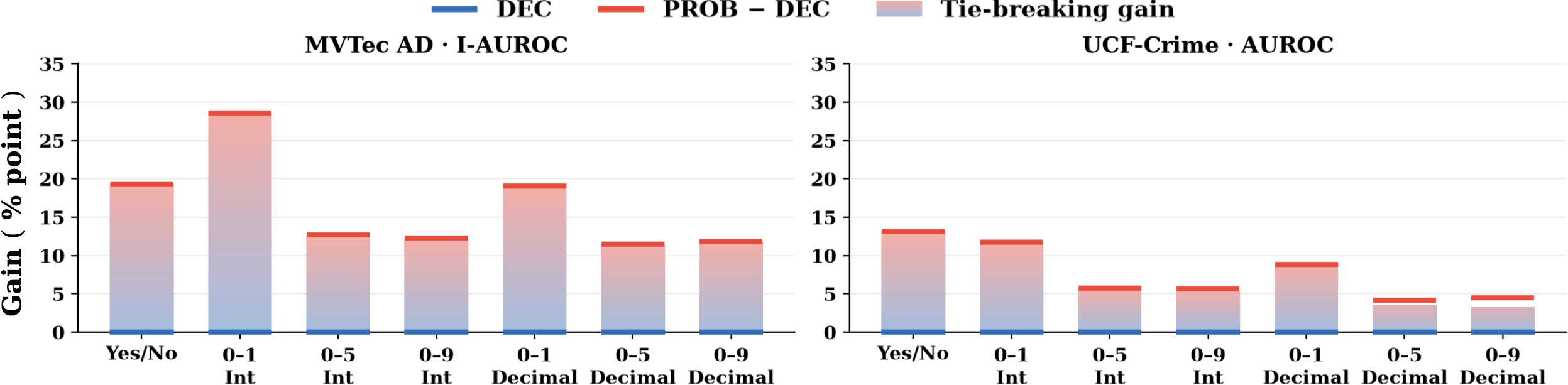}
\caption{
    Breaking \DEC ties recovers nearly all of the \PROB--\DEC performance gap.
    For each answer scale, the blue line marks the \DEC baseline at zero gain,
    the red line marks the full \PROB--\DEC gain, and the shaded bar marks the tie-breaking gain.
    Results are averaged over four VLMs.
}
\label{fig:rank_compression}
\vspace{-1em}
\end{figure*}

\subsection{Rank Compression}
\label{sec:decomposition}
Where does the improvement come from?
Compared with \DEC, \PROB can change the ordering of examples with the same decoded answer or
of examples with different decoded answers.

\paragraph{Breaking decoded-answer ties recovers nearly all of the gap.}
To isolate the first effect, 
we preserve \DEC's ordering between different decoded answers.
For examples with the same decoded answer, 
we break ties by sorting them in descending order of their \PROB scores.
Under Yes/No, all examples decoded as ``Yes'' remain above those decoded as ``No'', 
while examples within each group are ordered by \PROB.

\Cref{fig:rank_compression} shows how much of the full \PROB-\DEC performance gap is 
recovered by breaking decoded-answer ties.
For the two binary answer scales, tie-breaking reproduces the \PROB ranking by construction.
On the five finer scales,
\PROB can also change the order between examples with different decoded answers.
Even so, tie-breaking recovers at least 90.7\% of the average
primary-metric gap across these scales on every benchmark.
Across all seven scales, it recovers at least 95.2\% of the average
gap on every benchmark.
Thus, decoded-answer ties explain nearly all of the average
performance gap.
Full results are reported in \Cref{sec:supp_decomp}.

\begin{figure*}[t]
\centering
\includegraphics[width=\textwidth]{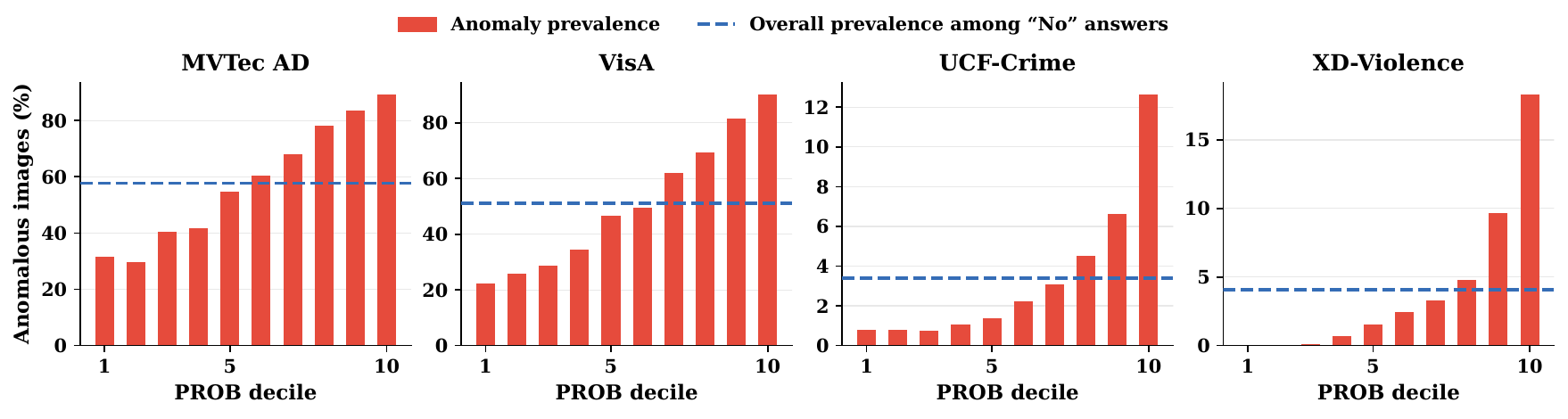}
\caption{
    Answer probabilities preserve useful order among inputs decoded as ``No''.
    For each VLM, we sort the ``No'' inputs by \PROB and divide them into deciles.
    Bars show anomaly prevalence within each decile, averaged over four VLMs,
    and the dashed line shows the corresponding prevalence over all ``No'' inputs.
    Higher-\PROB deciles contain more anomalies on all four benchmarks.
}
\label{fig:within_answer}
\end{figure*}

\paragraph{Answer probabilities preserve useful order within decoded-answer ties.}
To test whether \PROB preserves useful information within a decoded-answer tie, 
we focus on examples decoded as ``No'' under Yes/No.
\DEC assigns all of these examples the same score.
We sort them by \PROB and divide them into ten equally sized groups (deciles).
\Cref{fig:within_answer} shows that higher-\PROB groups contain a larger fraction of anomalies on all four benchmarks.

To quantify this ordering, we compute AUROC within the ``No'' group for each VLM.
On every benchmark, all VLMs achieve above-chance, 
with values ranging from 71.3 to 89.4 (\Cref{tab:within_answer}).
Thus, answer probabilities preserve useful ordering information even among examples decoded as ``No''.

\paragraph{Even an oracle binary threshold does not close the gap.}
One possible explanation for the \PROB--\DEC gap is that decoded scoring uses poorly placed decision boundaries rather than discarding useful ordering information.
We test this explanation in the binary setting, where \DEC is exactly obtained by thresholding \PROB at 0.5.
For each VLM and benchmark, we use the test labels to choose the threshold that maximizes the primary metric of the resulting binary scores.
This oracle threshold improves binary scoring by 0.89--27.58 points, but a residual gap of 0.41--17.31 points remains across all 16 model--benchmark settings (\Cref{tab:oracle_threshold}).
Thus, in the binary setting, poor threshold placement explains part of the performance loss but not the full \PROB--\DEC gap.

\begin{figure*}[t]
\centering
\includegraphics[width=\textwidth]{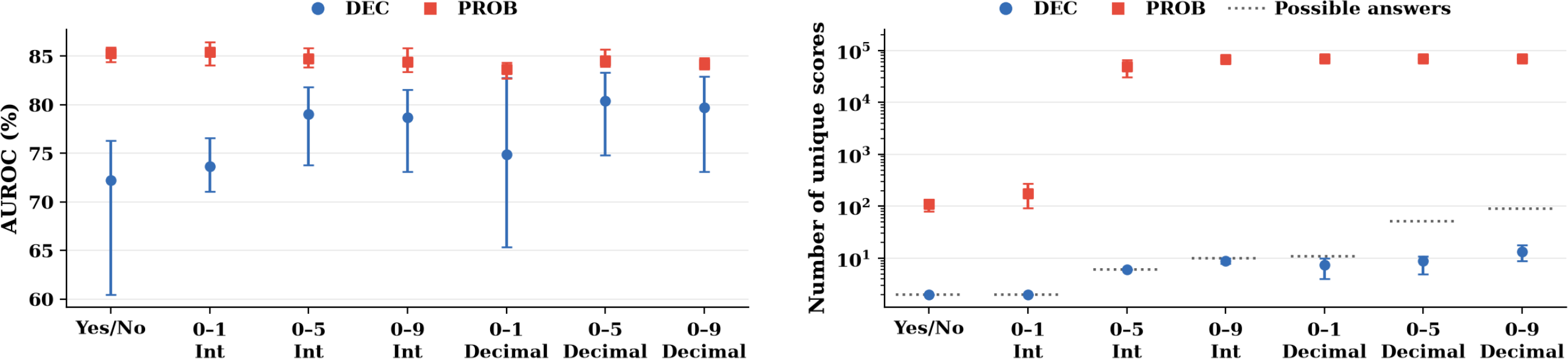}
\caption{
Answer-scale performance and number of distinct scores on UCF-Crime.
Left: frame-level AUROC across the seven answer scales.
Right: the number of distinct scores across the 69,634 UCF-Crime segments,
shown on a logarithmic scale.
Dotted gray lines indicate the number of possible answers
$|\mathcal{A}_k|$.
Markers report the mean over four VLMs, and vertical bars span the minimum and maximum across models.
}
\label{fig:answer_scale_cardinality}
\end{figure*}

\subsection{Answer Scales}
\label{sec:scales}
If decoded-answer ties account for most of the performance gap, 
a natural way to reduce rank compression is to use a finer answer scale.
We therefore compare seven answer scales ranging from 2 to 91 possible answers.
We ask whether finer scales improve \DEC and narrow the \PROB--\DEC gap, 
and whether the VLMs actually use the additional answer choices.

\paragraph{Finer answer scales help, but most additional answers are not used.}
Averaged over the four VLMs, all five nonbinary scales improve
\DEC over Yes/No and narrow the \PROB--\DEC gap on every
benchmark's primary metric (\Cref{tab:raw_summary}).
\Cref{fig:answer_scale_cardinality} illustrates this behavior on UCF-Crime and shows how the number of distinct scores changes across answer scales.
The improvement in \DEC is not monotonic in the number of possible answers.
To understand why, we examine how many distinct \DEC scores the models actually produce.
Under the 6- and 10-answer integer scales, \DEC uses all or nearly all possible answers.
The decimal scales allow 11, 51, and 91 possible answers, but \DEC produces only 7.5, 8.8, and 13.5 distinct scores on average, respectively.
In contrast, \PROB produces 69,538--69,542 distinct scores for the 69,634 UCF-Crime segments under the same decimal scales.
The other three benchmarks show the same pattern (\Cref{app:scale}).
Thus, increasing the number of possible answers does not translate into a comparable increase in the number of distinct \DEC scores.

\paragraph{More distinct scores do not imply better ranking.}
The previous result does not imply that producing more distinct scores is itself sufficient for better ranking performance.
We test this using the remaining ties in binary \PROB scores, comparing bfloat16 and float32 runs of Qwen3-VL-8B with Yes/No.
On UCF-Crime, the number of distinct segment scores increases from 137 to 65,885, while AUROC changes only from 85.26 to 85.27.
On MVTec AD, the mean number of distinct scores per category increases from 53.1 to 115.0, while I-AUROC changes from 91.51 to 91.78.
Thus, the number of distinct scores alone does not determine how well those scores rank anomalous examples above normal ones (\Cref{sec:supp_precision}).

\subsection{Alternative Scoring Strategies}
\label{sec:alternatives}
Can the remaining \PROB--\DEC gap be removed by changing other choices in how the VLM output is converted into an anomaly score?
We test alternative answer selection and aggregation rules, temporal smoothing, and prompt and output variations.

\paragraph{Alternative answer selection does not close the gap.}
We first ask whether the gap is specific to selecting the most probable answer.
Using the saved Yes/No answer distributions, we simulate temperature sampling, nucleus sampling, and majority voting on both video benchmarks.
Each strategy still reduces the model output to a single binary score.
Averaged over the four VLMs, none outperforms \DEC on either benchmark, and all remain below \PROB.
For example, the best tested strategy on UCF-Crime reaches 72.13 AUROC, compared with 72.20 for \DEC and 85.32 for \PROB.

\begin{figure*}[t]
\centering
\includegraphics[width=\textwidth]{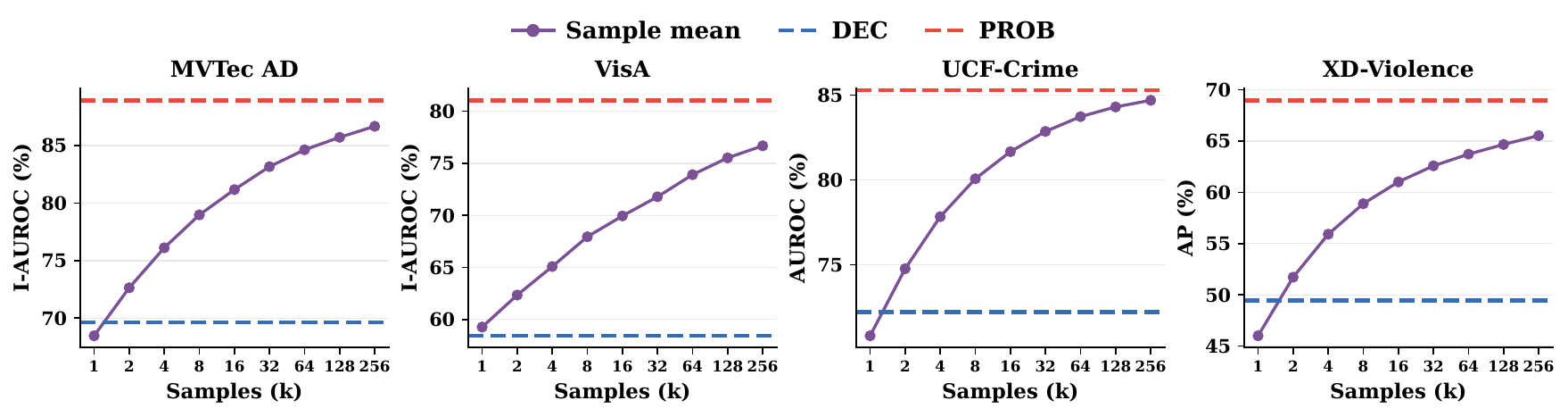}
\caption{
Averaging sampled answers narrows the gap to \PROB on all four benchmarks.
For each input, the sample mean is the fraction of Yes answers among
$k$ draws from the saved Yes/No answer distribution.
The sample count $k$ doubles from 1 to 256.
Each point averages the benchmark metric over three sampling seeds for each VLM and then over four VLMs.
Dashed lines show the means over four VLMs for \DEC and \PROB.
}
\label{fig:sampled_answers}
\end{figure*}

We then average the sampled answer values instead of selecting a single answer.
For Yes/No, the fraction of ``Yes'' answers among $k$ samples estimates the probability of ``Yes'', which \PROB uses directly.
\Cref{fig:sampled_answers} shows that performance improves with the number of samples on all four benchmarks.
On MVTec AD, mean I-AUROC increases from 81.18 with 16 samples to 86.68 with 256 samples, compared with 88.94 for \PROB.
Even with 256 samples, mean performance remains below \PROB on every benchmark (\Cref{sec:supp_sampling}).
Thus, changing which single answer is selected does not recover the lost ordering, while averaging multiple answers progressively approaches probability-weighted scoring.

\paragraph{Temporal smoothing narrows but does not close the mean gap.}
Temporal smoothing can reduce decoded-score ties by averaging scores from neighboring video segments.
We apply the same Gaussian smoothing to \DEC and \PROB within each video, using $\sigma\in\{1,\ldots,9\}$, and average results over the four VLMs and seven answer scales.
As smoothing increases, \DEC produces more distinct scores and the mean \PROB--\DEC gap narrows.
At $\sigma=9$, the mean number of distinct \DEC scores on UCF-Crime increases from 7.0 to about 31,745, while the mean AUROC gap falls from 7.66 to 2.28 points.
On XD-Violence, the mean AP gap falls from 10.97 to 3.01 points.
\PROB nevertheless remains higher on average at every tested value of $\sigma$ on both benchmarks (\Cref{sec:supp_sigma}).

\paragraph{Prompt and output variations do not reverse the comparison.}
We finally test whether the \PROB--\DEC comparison depends on the base prompt or output format.
Across three Yes/No paraphrases with Qwen3-VL-8B and InternVL3.5-8B on MVTec AD and UCF-Crime, \PROB remains higher on both metrics in every comparison (\Cref{sec:supp_prompt_sens}).
The same holds when we reverse the question and ask whether the example is normal, so that ``No'' indicates an anomaly (\Cref{sec:supp_polarity}).
We also add a brief factual description before or after the answer, using Qwen3-VL-8B and MiniCPM-V-4.5 on both binary scales across all four benchmarks(\Cref{sec:supp_explain}).
When the description comes first, both scoring rules use the answer distribution conditioned on the same description.
These changes affect absolute performance, but \PROB remains higher in all comparisons across both output orders.

\section{How Ties Affect Evaluation}
\label{sec:ties}
Decoded-answer scoring often assigns the same anomaly score to many examples.
This raises two evaluation choices: 
how equal scores are handled and 
how area is computed from precision--recall points.
We examine these two choices.

\paragraph{Evaluate equal scores together.}
When multiple frames receive the same anomaly score,
the detector provides no ordering among them.
As a score threshold passes that value,
all tied frames should be included together.
The UCF-Crime evaluation instead processes tied frames one at a time
and updates the ROC curve after each frame
(\Cref{sec:supp_official_metrics}).
This treats tied frames as if they had an order,
even though their scores do not distinguish them.
It also creates intermediate ROC points that do not correspond to any score threshold.
Because these points depend on the order within each tie,
the resulting ROC area can also depend on that order
~\citep{fawcett2006introduction}.

Let $\tau$ denote the fraction of anomalous--normal pairs that receive the same score.
Let $\mathrm{AUROC}^{-}$ and $\mathrm{AUROC}^{+}$ denote the smallest and largest
item-wise ROC areas obtainable by ordering frames within ties.
Relative to the half-credit AUROC in \Cref{eq:auc}, these extrema are
\begin{equation}
    \mathrm{AUROC}^{-}
    =
    \mathrm{AUROC}-\frac{\tau}{2},
    \qquad
    \mathrm{AUROC}^{+}
    =
    \mathrm{AUROC}+\frac{\tau}{2}.
    \label{eq:auc_order}
\end{equation}
The difference between these extrema is therefore exactly $\tau$.
Averaged over the four VLMs and seven answer scales on UCF-Crime,
this difference is 31.32 percentage points for \DEC and 0.23 points for \PROB
(\Cref{tab:ap_tie_sensitivity}).
We therefore compute AUROC by evaluating equal scores together,
using the half-credit definition in \Cref{eq:auc}.

\begin{figure*}[t]
\centering
\includegraphics[width=\textwidth]{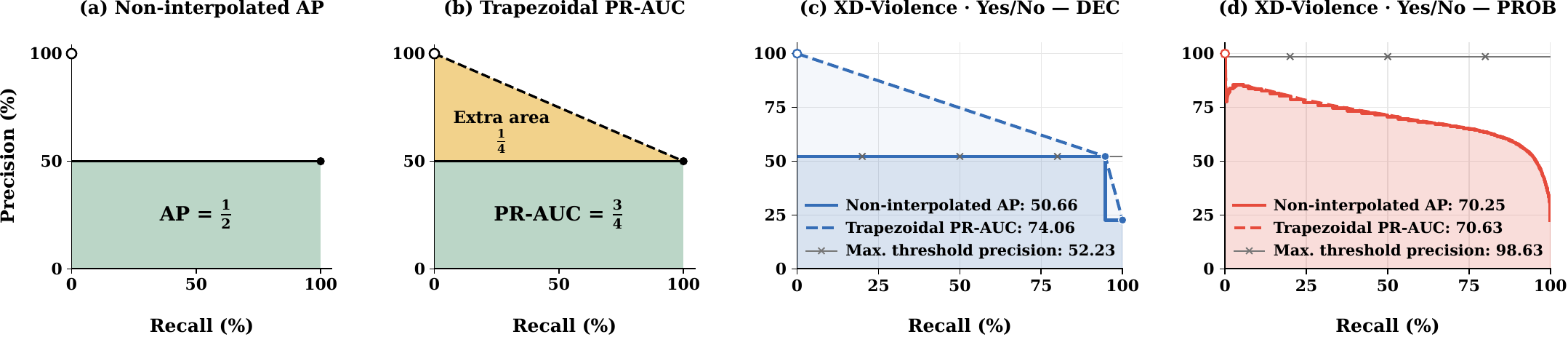}
\caption{
(a,b) Schematic comparison of non-interpolated AP and trapezoidal PR-AUC calculations.
(c,d) Qwen3-VL-8B with two scoring methods on XD-Violence.
Both calculations evaluate equal scores together.
Open circles mark the conventional endpoint at zero recall.
Gray lines mark maximum precision across thresholds that predict at least
one anomaly.
}
\label{fig:pr_area_example}
\end{figure*}

\paragraph{Linear PR interpolation can inflate reported performance.}
Even when equal scores are evaluated together, 
AP and trapezoidal PR-AUC can differ on the same precision--recall points.
Let $q_j=\mathrm{TP}_j/(\mathrm{TP}_j+\mathrm{FP}_j)$ denote precision at threshold $u_j$.
In \Cref{eq:ap}, 
the recall increase at this threshold is $p_j/P$, and 
AP uses $q_j$ for this increase.
Trapezoidal PR-AUC instead averages $q_{j-1}$ and $q_j$, 
with the conventional endpoint $q_0=1$ at zero recall~\citep{davis2006relationship,chen2024auprc}.
The difference between the two calculations is
\begin{equation}
    \mathrm{PR\mbox{-}AUC}_{\mathrm{trapezoids}}(s)
    -
    \operatorname{AP}(s)
    =
    \frac{1}{2}
    \sum_{j=1}^{J}
    \frac{p_j}{P}
    (q_{j-1}-q_j).
    \label{eq:pr_area_comparison}
\end{equation}
\Cref{fig:pr_area_example}(a,b) illustrates the corresponding rectangles and trapezoids.
When precision drops as recall increases, the trapezoid adds area above the AP rectangle, with more added area for larger recall increases and larger precision drops.

\Cref{fig:pr_area_example}(c,d) shows a concrete example for Qwen3-VL-8B with Yes/No on XD-Violence.
For \DEC, no threshold that predicts at least one frame as anomalous attains precision above 52.23\%.
Yet its trapezoidal PR-AUC is 74.06\%, compared with an AP of 50.66\%.
The line from the conventional $(0,1)$ endpoint to the first measured PR point spans a large recall increase and includes precision values above this attainable maximum.
For \PROB, AP and trapezoidal PR-AUC are 70.25\% and 70.63\%, respectively.
Thus, AP favors \PROB, whereas trapezoidal PR-AUC favors \DEC, without changing either set of anomaly scores.

This reversal is not specific to this example.
Holding the anomaly scores fixed, we compare AP and trapezoidal PR-AUC
for each answer scale, averaging over the four VLMs.
On both image benchmarks, trapezoidal PR-AUC narrows the mean
\PROB--\DEC difference, but \PROB remains higher under every scale.
On the video benchmarks, it reverses the mean comparison under
four of seven scales on UCF-Crime and six of seven on XD-Violence
(\Cref{tab:ap_estimator}).

\section{Discussion and Limitations}
\label{sec:limitations}

\paragraph{Decoded-answer performance can understate useful anomaly information.}
An anomaly score derived from a decoded answer does not necessarily reflect all of the useful ordering information in the VLM output.
Our results show that examples receiving the same decoded answer can still differ systematically in their answer probabilities.
Poor performance with decoded-answer scoring should therefore not, by itself, 
be interpreted as evidence that the VLM provides little useful information for anomaly detection.

\paragraph{Tied scores make evaluation choices consequential.}
When examples receive the same anomaly score,
different evaluation conventions can assign different credit without any change in the scores.
Processing tied scores one at a time introduces an ordering that the scores do not specify and can change AUROC.
Linear interpolation in precision--recall space can likewise add area between threshold points and, in our experiments, even reverse method comparisons.
These differences arise from the evaluation procedure rather than from changes in the anomaly scores themselves.
Evaluation results should therefore state how ties are handled and distinguish non-interpolated AP from trapezoidal PR-AUC.

\paragraph{Access to answer probabilities.}
Probability-weighted scoring requires the probabilities assigned to the possible answers.
Our experiments use open VLMs for which we can compute these probabilities.
We do not evaluate proprietary model APIs, where the available model outputs may differ.
Our sampled-answer experiments draw from the answer distributions computed for the open VLMs and therefore do not evaluate repeated sampling through such APIs.

\paragraph{Comparisons between answer scales.}
Within an answer scale, \DEC and \PROB operate on the same answer distribution and differ only in how it is converted into an anomaly score.
Across answer scales, however, the prompt, answer format, and meaning of numerical answers can also change.
Our comparisons across scales therefore measure the combined effect of these changes, not the effect of the number of possible answers alone.

\section{Conclusion}
We show that decoded-answer rank compression can substantially degrade anomaly ranking.
Across the image and video benchmarks we evaluate, decoded-answer scoring consistently yields lower AUROC and AP than probability-weighted scoring.
Using answer probabilities only to break decoded-answer ties
recovers at least 95\% of the average primary-metric gap on
every benchmark.
We also show that trapezoidal PR-AUC can mask this ranking loss
and even reverse the comparison between \DEC and \PROB without
changing the anomaly scores.
Evaluations should group equal scores and distinguish
non-interpolated AP from trapezoidal PR-AUC.

These results show that converting a VLM output into an anomaly score 
is not a neutral post-processing step.
This conversion can discard useful ordering information, 
and evaluation choices can further change how that loss appears in reported performance.
Both score construction and evaluation are therefore important 
when interpreting the performance of VLM-based anomaly detectors.
\newpage

\section*{AI Use Statement}

Generative AI tools were used to assist with sentence editing, grammar checking, typographical checks, and isolated utility functions in the inference code.
The authors reviewed every AI-assisted change, 
verified the mathematical definitions and all reported values against the evaluation outputs, and 
take responsibility for the final text, claims, code, and artifacts.
No reported experimental result was generated or selected by a language model.

\section*{Ethics Statement}

This work evaluates scoring rules on public image and video anomaly-detection benchmarks and redistributes no benchmark data.
MVTec AD and VisA depict industrial objects and textures.
UCF-Crime and XD-Violence include real events involving people.
We use them only to evaluate aggregate frame-level rankings and do not identify, profile, or characterize individuals.
We do not release generated descriptions of identifiable people.
As with any anomaly detector, deployment in industrial inspection or surveillance requires domain validation, human oversight, and safeguards against false alarms and misuse.

\section*{Reproducibility Statement}

Every main result uses frozen public checkpoints and deterministic scoring, with no training and, outside the explicitly seeded sampling controls, no stochastic decoding.
The checkpoints are \texttt{Qwen/Qwen3-VL-8B-Instruct}, \texttt{Qwen/Qwen2.5-VL-7B-Instruct}, \texttt{OpenGVLab/InternVL3\_5-8B}, and \texttt{openbmb/MiniCPM-V-4\_5}.
Each image is scored independently and image metrics are computed within category before macro-averaging.
For video, non-overlapping 16-frame units are judged from eight frames sampled uniformly over a centered 300-frame window and expanded back to the annotation grid (\Cref{sec:supp_protocol}).
The released artifact will include the exact domain-specific prompts, answer strings, launch commands, and evaluation scripts.
\Cref{sec:supp_prompts} documents the prompt structure and all currently recorded variants.

Both scores are computed from one set of answer sequence likelihoods per visual unit.
Candidate likelihoods are evaluated without special tokens; multi-token candidates use the sum of their token log-likelihoods.
Evaluation uses the grouped-threshold AP of \Cref{eq:ap} and tie-aware AUROC of \Cref{eq:auc}.
Video bootstrap replicates resample whole videos; image replicates resample normal and anomalous images separately within category; every replicate stream is shared across the paired scores, models, and scales.
Code and exact commands for every table will be released.

\bibliography{iclr2027_conference}
\bibliographystyle{iclr2027_conference}

\newpage

\appendix
\section{Experimental Details}
\label{app:likelihoods}
\label{app:protocol}

\subsection{Inputs and scoring computation}
\label{sec:supp_protocol}
We use the frozen checkpoints \texttt{Qwen/Qwen3-VL-8B-Instruct}, \texttt{Qwen/Qwen2.5-VL-7B-Instruct}, \texttt{OpenGVLab/InternVL3\_5-8B}, and \texttt{openbmb/MiniCPM-V-4\_5}.
All main runs use bfloat16 inference without temporal smoothing.

For images, we use 1,725 MVTec AD test images across 15 categories (467 normal and 1,258 anomalous) and 2,162 VisA test images across 12 categories (962 normal and 1,200 anomalous).
MVTec AD uses the category test directories with \texttt{good} labeled normal.
VisA uses test rows in \nolinkurl{split_csv/1cls.csv}.
Each input is EXIF-oriented, converted to RGB, and passed to the model's image processor as a single image.
The prompt includes the category name but no labels, defect subtypes, paths, normal-reference images, or pixel masks.
Image AUROC and AP are computed within category and macro-averaged.

For videos, the UCF-Crime and XD-Violence test splits contain 290 and 800 videos.
We score 1,111,808 and 2,335,801 frames, including 88,094 and 529,797 anomalous frames, respectively.
Non-overlapping 16-frame segments give 69,634 and 146,449 predictions.
Each segment uses eight frames sampled uniformly over a centered 300-frame window clipped at video boundaries.
Qwen3-VL, Qwen2.5-VL, and InternVL3.5 receive an ordered image sequence.
MiniCPM-V-4.5 receives the same frames in a $4\times2$ contact sheet.
We repeat each segment score over its frames and concatenate videos before evaluation.

Single-token answers use logits at the answer position.
Multi-token answers use teacher-forced continuation scoring, 
summing token log-probabilities without adding special tokens.
We normalize these full-sequence probabilities over the answer set at $T=1$ and compute both scores from the same distribution.
The main comparison does not sample answers.

\paragraph{Agreement with unconstrained decoding.}
\label{sec:supp_admissible}
For Qwen3-VL-8B and InternVL3.5-8B with Yes/No, the unconstrained top-1 token agrees with \DEC on every UCF-Crime segment and MVTec AD image.
Mean probability mass on Yes/No is 0.9999999 and 0.9999750 on UCF-Crime, respectively.
Minimum mass exceeds 0.99978 on MVTec AD.
This check validates the constrained binary answer selection for these two models and benchmarks.

\subsection{Prompts and answer scales}
\label{sec:supp_prompts}
The following templates instantiate the seven answer scales in \Cref{tab:answer_scales}.
Integer scales use $M\in\{1,5,9\}$ and decimal scales use $M\in\{1.0,5.0,9.0\}$.
In image prompts, \texttt{<category>} replaces underscores with spaces and inserts a space before a trailing numeric suffix, as in \texttt{metal nut} and \texttt{macaroni 1}.
The article ``a'' is retained for every category.

\paragraph{Image Yes/No.}
\begin{quote}\small\ttfamily\raggedright
You are inspecting an image of a <category>.\\
Assess whether the object or material is anomalous or defective, for example damaged, contaminated, malformed, misplaced, or missing a part.\\
Answer with exactly one word: Yes or No.
\end{quote}

\paragraph{Image integer scales.}
\begin{quote}\small\ttfamily\raggedright
You are inspecting an image of a <category>.\\
Assess whether the object or material is anomalous or defective, for example damaged, contaminated, malformed, misplaced, or missing a part.\\
Rate the anomaly severity from 0 to $M$.\\
0 means completely normal.\\
$M$ means clearly anomalous or defective.\\
Answer with exactly one integer from 0 to $M$ and no other text.
\end{quote}

\paragraph{Image one-decimal scales.}
\begin{quote}\small\ttfamily\raggedright
You are inspecting an image of a <category>.\\
Assess whether the object or material is anomalous or defective, for example damaged, contaminated, malformed, misplaced, or missing a part.\\
Rate the anomaly severity from 0.0 to $M$.\\
0.0 means completely normal.\\
$M$ means clearly anomalous or defective.\\
Answer with exactly one float rounded to one decimal place from 0.0 to $M$ and no other text.
\end{quote}

The image base prompt version is \texttt{industrial-query-v1}.
The decoder-agreement and float32 controls reuse its Yes/No wording.

\paragraph{Video Yes/No.}
\begin{quote}\small\ttfamily\raggedright
You are reviewing ordered frames from a surveillance video clip.\\
Is the clip anomalous, suspicious, violent, criminal, or dangerous?\\
Answer with exactly one word: Yes or No.
\end{quote}

\paragraph{Video integer scales.}
\begin{quote}\small\ttfamily\raggedright
You are reviewing ordered frames from a surveillance video clip.\\
Rate how anomalous, suspicious, violent, criminal, or dangerous the clip is.\\
Use a single integer from 0 to $M$:\\
0 means completely normal.\\
$M$ means clearly anomalous or dangerous.\\
Answer with exactly one integer from 0 to $M$ and no other text.
\end{quote}

\paragraph{Video one-decimal scales.}
\begin{quote}\small\ttfamily\raggedright
You are reviewing ordered frames from a surveillance video clip.\\
Rate how anomalous, suspicious, violent, criminal, or dangerous the clip is on a scale from 0.0 to $M$.\\
0.0 means completely normal.\\
$M$ means clearly anomalous or dangerous.\\
Answer with exactly one float rounded to one decimal place from 0.0 to $M$ and no other text.
\end{quote}

\subsection{Paired uncertainty estimates}
\label{sec:supp_bootstrap}
We use 10,000 nonparametric bootstrap replicates shared across scoring rules, models, and answer scales.
Video replicates resample complete videos with replacement and recompute frame-level metrics.
Image replicates keep categories fixed, resample normal and anomalous images separately within each category, and macro-average the recomputed metrics.
The image seed is 20260903 with deterministic dataset/category offsets.
The intervals measure test-set sampling uncertainty conditional on the evaluated models and categories.
\Cref{tab:bootstrap_scales} reports the primary-metric intervals.

\section{Complete Results}
\label{app:complete_results}

\subsection{Model and answer-scale comparisons}
\label{sec:supp_raw}
\Cref{tab:image_raw_mvtec,tab:image_raw_visa,tab:raw_ucf,tab:raw_xd} report both metrics for every main model--scale--benchmark configuration.
All 224 paired differences favor \PROB.
\Cref{tab:bootstrap_scales} gives the corresponding scale-averaged primary-metric gains and confidence intervals.

\begin{table*}[tbp]
\centering
\caption{Category-macro image-level performance on MVTec AD for every model and answer scale (\%). Each category is weighted equally. \PROB exceeds \DEC in every cell.}
\label{tab:image_raw_mvtec}
\setlength{\tabcolsep}{2.5pt}
\small
\resizebox{\textwidth}{!}{%
\begin{tabular}{llcccccccccccccc}
\toprule
& & \multicolumn{2}{c}{Yes/No} & \multicolumn{2}{c}{0--1 int} & \multicolumn{2}{c}{0--5 int} & \multicolumn{2}{c}{0--9 int} & \multicolumn{2}{c}{0--1 float} & \multicolumn{2}{c}{0--5 float} & \multicolumn{2}{c}{0--9 float} \\
\cmidrule(lr){3-4}\cmidrule(lr){5-6}\cmidrule(lr){7-8}\cmidrule(lr){9-10}\cmidrule(lr){11-12}\cmidrule(lr){13-14}\cmidrule(lr){15-16}
Metric & Model & \DEC & \PROB & \DEC & \PROB & \DEC & \PROB & \DEC & \PROB & \DEC & \PROB & \DEC & \PROB & \DEC & \PROB \\
\midrule
\multirow{5}{*}{I-AUROC}
& Qwen3-VL-8B      & 75.05 & \underline{91.51} & 67.96 & \underline{91.46} & 78.59 & \underline{90.69} & 78.66 & \underline{90.76} & 82.08 & \underline{91.11} & 80.03 & \underline{90.90} & 78.56 & \underline{90.81} \\
& Qwen2.5-VL-7B    & 59.26 & \underline{90.82} & 52.05 & \underline{90.29} & 81.06 & \underline{91.39} & 79.79 & \underline{91.17} & 73.81 & \underline{91.28} & 81.23 & \underline{91.37} & 79.64 & \underline{91.52} \\
& InternVL3.5-8B   & 69.47 & \underline{84.18} & 51.94 & \underline{83.74} & 69.84 & \underline{84.29} & 71.26 & \underline{83.77} & 70.53 & \underline{84.74} & 70.76 & \underline{84.12} & 72.11 & \underline{83.62} \\
& MiniCPM-V-4.5-8B & 74.73 & \underline{89.25} & 72.20 & \underline{93.18} & 78.79 & \underline{92.88} & 79.07 & \underline{92.21} & 56.46 & \underline{92.03} & 81.18 & \underline{92.57} & 81.38 & \underline{92.85} \\
\cmidrule(lr){2-16}
& Mean             & 69.63 & \textbf{88.94} & 61.04 & \textbf{89.67} & 77.07 & \textbf{89.81} & 77.19 & \textbf{89.48} & 70.72 & \textbf{89.79} & 78.30 & \textbf{89.74} & 77.92 & \textbf{89.70} \\
\midrule
\multirow{5}{*}{I-AP}
& Qwen3-VL-8B      & 85.56 & \underline{95.67} & 81.87 & \underline{95.55} & 87.62 & \underline{95.38} & 87.47 & \underline{95.52} & 89.16 & \underline{95.75} & 88.43 & \underline{95.48} & 87.84 & \underline{95.51} \\
& Qwen2.5-VL-7B    & 77.44 & \underline{95.56} & 73.44 & \underline{95.30} & 88.27 & \underline{95.88} & 87.61 & \underline{95.77} & 84.58 & \underline{95.83} & 88.48 & \underline{95.92} & 87.81 & \underline{95.93} \\
& InternVL3.5-8B   & 82.16 & \underline{92.11} & 73.41 & \underline{91.90} & 82.56 & \underline{92.55} & 83.22 & \underline{92.11} & 83.12 & \underline{92.62} & 83.36 & \underline{92.30} & 84.13 & \underline{91.80} \\
& MiniCPM-V-4.5-8B & 85.51 & \underline{95.52} & 84.30 & \underline{96.98} & 87.47 & \underline{96.88} & 87.54 & \underline{96.62} & 76.04 & \underline{96.88} & 88.92 & \underline{96.89} & 89.11 & \underline{96.97} \\
\cmidrule(lr){2-16}
& Mean             & 82.67 & \textbf{94.71} & 78.25 & \textbf{94.93} & 86.48 & \textbf{95.17} & 86.46 & \textbf{95.01} & 83.22 & \textbf{95.27} & 87.30 & \textbf{95.15} & 87.22 & \textbf{95.05} \\
\bottomrule
\end{tabular}%
}
\end{table*}

\begin{table*}[tbp]
\centering
\caption{Category-macro image-level performance on VisA for every model and answer scale (\%). Each category is weighted equally. \PROB exceeds \DEC in every cell.}
\label{tab:image_raw_visa}
\setlength{\tabcolsep}{2.5pt}
\small
\resizebox{\textwidth}{!}{%
\begin{tabular}{llcccccccccccccc}
\toprule
& & \multicolumn{2}{c}{Yes/No} & \multicolumn{2}{c}{0--1 int} & \multicolumn{2}{c}{0--5 int} & \multicolumn{2}{c}{0--9 int} & \multicolumn{2}{c}{0--1 float} & \multicolumn{2}{c}{0--5 float} & \multicolumn{2}{c}{0--9 float} \\
\cmidrule(lr){3-4}\cmidrule(lr){5-6}\cmidrule(lr){7-8}\cmidrule(lr){9-10}\cmidrule(lr){11-12}\cmidrule(lr){13-14}\cmidrule(lr){15-16}
Metric & Model & \DEC & \PROB & \DEC & \PROB & \DEC & \PROB & \DEC & \PROB & \DEC & \PROB & \DEC & \PROB & \DEC & \PROB \\
\midrule
\multirow{5}{*}{I-AUROC}
& Qwen3-VL-8B      & 62.00 & \underline{87.23} & 58.04 & \underline{87.79} & 65.60 & \underline{87.89} & 66.24 & \underline{87.68} & 71.09 & \underline{87.61} & 65.04 & \underline{87.56} & 64.76 & \underline{87.30} \\
& Qwen2.5-VL-7B    & 53.96 & \underline{81.27} & 49.96 & \underline{78.74} & 71.64 & \underline{82.23} & 71.19 & \underline{81.98} & 64.31 & \underline{82.29} & 70.19 & \underline{81.89} & 68.83 & \underline{81.85} \\
& InternVL3.5-8B   & 58.42 & \underline{79.47} & 50.75 & \underline{79.52} & 63.00 & \underline{79.99} & 64.79 & \underline{79.67} & 63.97 & \underline{79.64} & 63.66 & \underline{77.90} & 63.74 & \underline{74.70} \\
& MiniCPM-V-4.5-8B & 59.38 & \underline{76.27} & 58.08 & \underline{85.50} & 65.91 & \underline{85.77} & 66.17 & \underline{85.84} & 51.92 & \underline{86.82} & 66.45 & \underline{86.13} & 68.78 & \underline{86.02} \\
\cmidrule(lr){2-16}
& Mean             & 58.44 & \textbf{81.06} & 54.21 & \textbf{82.89} & 66.54 & \textbf{83.97} & 67.10 & \textbf{83.79} & 62.82 & \textbf{84.09} & 66.33 & \textbf{83.37} & 66.53 & \textbf{82.47} \\
\midrule
\multirow{5}{*}{I-AP}
& Qwen3-VL-8B      & 66.40 & \underline{89.12} & 62.88 & \underline{89.28} & 69.18 & \underline{89.81} & 69.59 & \underline{89.73} & 73.27 & \underline{89.71} & 68.61 & \underline{89.45} & 68.32 & \underline{89.39} \\
& Qwen2.5-VL-7B    & 59.84 & \underline{84.08} & 56.58 & \underline{81.64} & 73.15 & \underline{85.20} & 72.94 & \underline{84.95} & 67.73 & \underline{85.29} & 71.95 & \underline{84.97} & 71.19 & \underline{85.09} \\
& InternVL3.5-8B   & 63.24 & \underline{83.07} & 57.08 & \underline{82.95} & 66.73 & \underline{83.58} & 68.04 & \underline{83.26} & 67.36 & \underline{83.32} & 67.56 & \underline{81.91} & 67.57 & \underline{80.01} \\
& MiniCPM-V-4.5-8B & 63.59 & \underline{81.14} & 62.62 & \underline{87.51} & 69.07 & \underline{87.99} & 69.02 & \underline{87.81} & 57.83 & \underline{88.39} & 69.62 & \underline{88.23} & 71.67 & \underline{88.28} \\
\cmidrule(lr){2-16}
& Mean             & 63.27 & \textbf{84.35} & 59.79 & \textbf{85.34} & 69.53 & \textbf{86.64} & 69.90 & \textbf{86.44} & 66.55 & \textbf{86.68} & 69.43 & \textbf{86.14} & 69.69 & \textbf{85.69} \\
\bottomrule
\end{tabular}%
}
\end{table*}

\begin{table*}[tbp]
\centering
\caption{
Frame-level performance on UCF-Crime for every model and answer scale
(\%). \PROB exceeds \DEC in every cell.
Underline marks the winner within each pair, and bold marks the winner in the
Mean row.
}
\label{tab:raw_ucf}
\setlength{\tabcolsep}{2.5pt}
\small
\resizebox{\textwidth}{!}{%
\begin{tabular}{llcccccccccccccc}
\toprule
& & \multicolumn{2}{c}{Yes/No} & \multicolumn{2}{c}{0--1 int} & \multicolumn{2}{c}{0--5 int} & \multicolumn{2}{c}{0--9 int} & \multicolumn{2}{c}{0--1 float} & \multicolumn{2}{c}{0--5 float} & \multicolumn{2}{c}{0--9 float} \\
\cmidrule(lr){3-4}\cmidrule(lr){5-6}\cmidrule(lr){7-8}\cmidrule(lr){9-10}\cmidrule(lr){11-12}\cmidrule(lr){13-14}\cmidrule(lr){15-16}
Metric & Model & \DEC & \PROB & \DEC & \PROB & \DEC & \PROB & \DEC & \PROB & \DEC & \PROB & \DEC & \PROB & \DEC & \PROB \\
\midrule
\multirow{5}{*}{AUROC}
& Qwen3-VL-8B      & 76.28 & \underline{85.26} & 75.23 & \underline{85.49} & 78.93 & \underline{84.43} & 78.97 & \underline{83.90} & 82.76 & \underline{84.31} & 82.85 & \underline{84.20} & 82.91 & \underline{83.90} \\
& Qwen2.5-VL-7B    & 60.47 & \underline{85.64} & 71.08 & \underline{85.61} & 73.78 & \underline{85.81} & 73.10 & \underline{85.87} & 69.88 & \underline{82.70} & 74.81 & \underline{84.20} & 73.11 & \underline{84.42} \\
& InternVL3.5-8B   & 75.97 & \underline{85.93} & 71.83 & \underline{86.46} & 81.79 & \underline{84.96} & 81.19 & \underline{84.34} & 81.65 & \underline{84.37} & 83.33 & \underline{85.73} & 81.68 & \underline{84.85} \\
& MiniCPM-V-4.5-8B & 76.07 & \underline{84.45} & 76.60 & \underline{84.10} & 81.70 & \underline{83.85} & 81.57 & \underline{83.40} & 65.34 & \underline{83.35} & 80.50 & \underline{83.94} & 81.23 & \underline{83.70} \\
\cmidrule(lr){2-16}
& Mean             & 72.20 & \textbf{85.32} & 73.69 & \textbf{85.42} & 79.05 & \textbf{84.76} & 78.71 & \textbf{84.38} & 74.91 & \textbf{83.68} & 80.37 & \textbf{84.52} & 79.73 & \textbf{84.22} \\
\midrule
\multirow{5}{*}{AP}
& Qwen3-VL-8B      & 19.91 & \underline{40.87} & 18.98 & \underline{39.85} & 26.47 & \underline{38.90} & 27.12 & \underline{37.16} & 30.52 & \underline{37.41} & 31.49 & \underline{37.34} & 31.86 & \underline{36.36} \\
& Qwen2.5-VL-7B    & 17.91 & \underline{37.46} & 23.13 & \underline{39.44} & 24.73 & \underline{39.79} & 23.99 & \underline{39.04} & 24.22 & \underline{38.46} & 26.49 & \underline{39.43} & 24.32 & \underline{39.11} \\
& InternVL3.5-8B   & 22.70 & \underline{39.39} & 23.39 & \underline{39.00} & 30.15 & \underline{39.67} & 29.45 & \underline{38.38} & 31.42 & \underline{39.67} & 31.25 & \underline{41.52} & 31.88 & \underline{39.72} \\
& MiniCPM-V-4.5-8B & 23.34 & \underline{41.79} & 20.08 & \underline{39.72} & 31.37 & \underline{40.34} & 26.54 & \underline{38.46} & 22.46 & \underline{37.09} & 31.15 & \underline{40.77} & 32.24 & \underline{39.42} \\
\cmidrule(lr){2-16}
& Mean             & 20.96 & \textbf{39.88} & 21.39 & \textbf{39.50} & 28.18 & \textbf{39.67} & 26.77 & \textbf{38.26} & 27.16 & \textbf{38.16} & 30.10 & \textbf{39.77} & 30.07 & \textbf{38.65} \\
\bottomrule
\end{tabular}%
}
\end{table*}

\begin{table*}[tbp]
\centering
\caption{
Frame-level performance on XD-Violence for every model and answer scale
(\%). \PROB exceeds \DEC in every cell.
Underline marks the winner within each pair, and bold marks the winner in the
Mean row.
}
\label{tab:raw_xd}
\setlength{\tabcolsep}{2.5pt}
\small
\resizebox{\textwidth}{!}{%
\begin{tabular}{llcccccccccccccc}
\toprule
& & \multicolumn{2}{c}{Yes/No} & \multicolumn{2}{c}{0--1 int} & \multicolumn{2}{c}{0--5 int} & \multicolumn{2}{c}{0--9 int} & \multicolumn{2}{c}{0--1 float} & \multicolumn{2}{c}{0--5 float} & \multicolumn{2}{c}{0--9 float} \\
\cmidrule(lr){3-4}\cmidrule(lr){5-6}\cmidrule(lr){7-8}\cmidrule(lr){9-10}\cmidrule(lr){11-12}\cmidrule(lr){13-14}\cmidrule(lr){15-16}
Metric & Model & \DEC & \PROB & \DEC & \PROB & \DEC & \PROB & \DEC & \PROB & \DEC & \PROB & \DEC & \PROB & \DEC & \PROB \\
\midrule
\multirow{5}{*}{AUROC}
& Qwen3-VL-8B      & 84.64 & \underline{91.40} & 80.35 & \underline{91.20} & 86.78 & \underline{90.17} & 87.16 & \underline{89.85} & 88.03 & \underline{89.30} & 88.74 & \underline{89.62} & 88.82 & \underline{89.66} \\
& Qwen2.5-VL-7B    & 64.33 & \underline{88.62} & 80.25 & \underline{88.87} & 83.45 & \underline{88.50} & 83.04 & \underline{88.70} & 80.59 & \underline{88.43} & 84.84 & \underline{88.36} & 84.83 & \underline{88.85} \\
& InternVL3.5-8B   & 85.31 & \underline{91.67} & 85.91 & \underline{91.59} & 89.15 & \underline{91.76} & 90.22 & \underline{91.73} & 89.07 & \underline{91.35} & 88.84 & \underline{91.45} & 89.38 & \underline{91.44} \\
& MiniCPM-V-4.5-8B & 84.92 & \underline{91.51} & 84.42 & \underline{91.20} & 88.33 & \underline{90.66} & 88.35 & \underline{90.73} & 79.85 & \underline{90.16} & 87.93 & \underline{90.85} & 88.41 & \underline{90.72} \\
\cmidrule(lr){2-16}
& Mean             & 79.80 & \textbf{90.80} & 82.73 & \textbf{90.71} & 86.93 & \textbf{90.27} & 87.20 & \textbf{90.25} & 84.39 & \textbf{89.81} & 87.59 & \textbf{90.07} & 87.86 & \textbf{90.17} \\
\midrule
\multirow{5}{*}{AP}
& Qwen3-VL-8B      & 50.66 & \underline{70.25} & 43.35 & \underline{69.42} & 54.60 & \underline{66.88} & 55.66 & \underline{66.01} & 57.64 & \underline{63.17} & 59.42 & \underline{63.58} & 59.84 & \underline{63.45} \\
& Qwen2.5-VL-7B    & 36.81 & \underline{59.19} & 51.13 & \underline{61.75} & 54.28 & \underline{62.11} & 53.91 & \underline{62.22} & 53.77 & \underline{63.22} & 56.28 & \underline{62.79} & 55.89 & \underline{64.07} \\
& InternVL3.5-8B   & 55.32 & \underline{73.31} & 56.61 & \underline{69.31} & 60.81 & \underline{69.13} & 64.37 & \underline{70.12} & 63.97 & \underline{69.49} & 60.88 & \underline{69.46} & 63.86 & \underline{70.87} \\
& MiniCPM-V-4.5-8B & 54.97 & \underline{73.17} & 51.01 & \underline{71.84} & 60.64 & \underline{69.73} & 60.91 & \underline{69.40} & 53.92 & \underline{66.80} & 61.06 & \underline{70.31} & 61.30 & \underline{69.03} \\
\cmidrule(lr){2-16}
& Mean             & 49.44 & \textbf{68.98} & 50.53 & \textbf{68.08} & 57.58 & \textbf{66.96} & 58.71 & \textbf{66.94} & 57.32 & \textbf{65.67} & 59.41 & \textbf{66.54} & 60.22 & \textbf{66.85} \\
\bottomrule
\end{tabular}%
}
\end{table*}

\begin{table}[tbp]
\centering
\caption{Primary-metric gains and paired 95\% bootstrap intervals in percentage points, using 10,000 replicates. Each scale averages four models. The final row averages 28 configurations.}
\label{tab:bootstrap_scales}
\small
\setlength{\tabcolsep}{3pt}
\begin{tabular}{lcccc}
\toprule
Scale & MVTec AD & VisA & UCF-Crime & XD-Violence \\
\midrule
Yes/No     & \shortstack{$+19.31$\\$[18.21,20.39]$} & \shortstack{$+22.62$\\$[21.53,23.71]$}  &  \shortstack{$+13.12$\\$[10.08,15.88]$} & \shortstack{$+19.54$\\$[17.61,21.40]$} \\
\addlinespace[2pt]
0--1 int   & \shortstack{$+28.63$\\$[27.58,29.65]$} & \shortstack{$+28.68$\\$[27.59,29.75]$}  &  \shortstack{$+11.73$\\$[9.25,14.36]$} & \shortstack{$+17.55$\\$[15.23,19.84]$} \\
\addlinespace[2pt]
0--5 int   & \shortstack{$+12.74$\\$[11.87,13.62]$} & \shortstack{$+17.43$\\$[16.45,18.42]$}  &  \shortstack{$+5.71$\\$[3.59, 7.87]$}  & \shortstack{$+9.38$\\$[7.39,11.47]$} \\
\addlinespace[2pt]
0--9 int   & \shortstack{$+12.28$\\$[11.40,13.17]$} & \shortstack{$+16.70$\\$[15.69,17.71]$}  &  \shortstack{$+5.67$\\$[4.01, 7.42]$}  & \shortstack{$+8.23$\\$[6.61, 9.89]$} \\
\addlinespace[2pt]
0--1 float & \shortstack{$+19.07$\\$[18.02,20.09]$} & \shortstack{$+21.27$\\$[20.26,22.25]$}  &  \shortstack{$+8.78$\\$[6.05,11.23]$}  & \shortstack{$+8.34$\\$[7.06, 9.60]$} \\
\addlinespace[2pt]
0--5 float & \shortstack{$+11.44$\\$[10.60,12.29]$} & \shortstack{$+17.04$\\$[15.99,18.09]$}  &  \shortstack{$+4.14$\\$[2.47, 5.89]$}  & \shortstack{$+7.12$\\$[5.53, 8.72]$} \\
\addlinespace[2pt]
0--9 float & \shortstack{$+11.77$\\$[10.91,12.64]$} & \shortstack{$+15.94$\\$[14.94,16.94]$}  &  \shortstack{$+4.49$\\$[2.71, 6.45]$}  & \shortstack{$+6.63$\\$[5.00, 8.25]$} \\
\addlinespace[2pt]
\midrule
All 28 pairs & \shortstack{$+16.46$\\$[15.67,17.26]$} & \shortstack{$+19.95$\\$[19.02,20.86]$}  &  \shortstack{$+7.66$\\$[5.56, 9.71]$} & \shortstack{$+10.97$\\$[9.37,12.58]$} \\
\addlinespace[2pt]
\bottomrule
\end{tabular}
\end{table}

All scale-averaged intervals exclude zero.
At the individual-configuration level, 28 of 28 XD-Violence AP intervals and 23 of 28 UCF-Crime AUROC intervals exclude zero.
The five UCF-Crime intervals that include zero are MiniCPM-V-4.5-8B at 0--5 integer ($+2.15\ [-0.28,+4.48]$) and 0--9 integer ($+1.83\ [-0.00,+3.19]$), and Qwen3-VL-8B at 0--1 decimal ($+1.55\ [-0.29,+2.98]$), 0--5 decimal ($+1.34\ [-0.56,+2.75]$), and 0--9 decimal ($+0.99\ [-0.92,+2.35]$).
All image per-model Yes/No intervals exclude zero.

\subsection{Image categories and video event classes}
\label{sec:supp_perclass}
Under Yes/No, \PROB improves both metrics in all 27 image categories and all 19 video event classes, averaged over four VLMs.
Each video class is evaluated with all normal test videos.
XD-Violence is multi-label, so a video contributes to each annotated event type and is compared with all 300 normal videos.
The per-class AP values reflect different anomaly prevalences.

\begin{table}[tbp]
\centering
\caption{Yes/No image results (\%), averaged over four models. Each category has equal weight within its benchmark.}
\label{tab:image_per_category}
\small
\setlength{\tabcolsep}{3pt}
\begin{tabular}{lrrrrrr}
\toprule
& \multicolumn{3}{c}{I-AUROC} & \multicolumn{3}{c}{I-AP} \\
Class/category & \DEC & \PROB & $\Delta$ & \DEC & \PROB & $\Delta$ \\
\midrule
\multicolumn{7}{l}{\emph{MVTec AD}} \\
Bottle     & 76.73 & 88.88 & +12.15 & 88.61 & 96.52 &  +7.91 \\
Cable      & 51.63 & 82.16 & +30.53 & 62.59 & 88.24 & +25.65 \\
Capsule    & 57.68 & 81.75 & +24.06 & 85.25 & 95.33 & +10.08 \\
Carpet     & 83.71 & 99.37 & +15.67 & 92.20 & 99.79 &  +7.59 \\
Grid       & 85.09 & 99.24 & +14.15 & 91.97 & 99.69 &  +7.72 \\
Hazelnut   & 73.35 & 92.28 & +18.93 & 79.96 & 95.51 & +15.55 \\
Leather    & 83.56 & 99.65 & +16.09 & 91.51 & 99.86 &  +8.35 \\
Metal nut  & 60.62 & 91.10 & +30.48 & 84.93 & 97.64 & +12.71 \\
Pill       & 64.36 & 76.99 & +12.63 & 88.90 & 94.99 &  +6.09 \\
Screw      & 56.51 & 86.89 & +30.38 & 77.71 & 94.67 & +16.96 \\
Tile       & 74.26 & 98.45 & +24.19 & 85.48 & 99.33 & +13.86 \\
Toothbrush & 74.38 & 77.78 &  +3.40 & 84.54 & 89.91 &  +5.37 \\
Transistor & 60.83 & 74.92 & +14.08 & 52.38 & 73.91 & +21.53 \\
Wood       & 77.50 & 99.68 & +22.18 & 89.18 & 99.87 & +10.70 \\
Zipper     & 64.24 & 84.97 & +20.73 & 84.76 & 95.41 & +10.65 \\
\midrule
\multicolumn{7}{l}{\emph{VisA}} \\
Candle      & 62.50 & 90.11 & +27.61 & 62.38 & 91.70 & +29.32 \\
Capsules    & 53.25 & 83.53 & +30.28 & 64.94 & 89.24 & +24.30 \\
Cashew      & 54.12 & 74.22 & +20.10 & 69.42 & 87.49 & +18.08 \\
Chewing gum & 76.00 & 96.86 & +20.86 & 84.00 & 98.51 & +14.51 \\
Fryum       & 65.50 & 91.02 & +25.52 & 77.00 & 95.37 & +18.37 \\
Macaroni 1  & 55.75 & 82.48 & +26.73 & 55.64 & 84.36 & +28.73 \\
Macaroni 2  & 55.50 & 67.08 & +11.58 & 53.80 & 64.83 & +11.02 \\
PCB 1       & 50.75 & 58.98 &  +8.23 & 50.75 & 62.72 & +11.97 \\
PCB 2       & 53.50 & 71.90 & +18.40 & 53.50 & 75.18 & +21.68 \\
PCB 3       & 50.88 & 69.86 & +18.99 & 50.63 & 71.46 & +20.83 \\
PCB 4       & 65.38 & 96.21 & +30.84 & 65.08 & 96.23 & +31.15 \\
Pipe fryum  & 58.13 & 90.46 & +32.34 & 72.08 & 95.16 & +23.08 \\
\bottomrule
\end{tabular}
\end{table}

\begin{table}[tbp]
\centering
\caption{Yes/No video results (\%), averaged over four models. Each event class includes all normal test videos. XD-Violence videos can contribute to multiple classes.}
\label{tab:video_per_class}
\small
\setlength{\tabcolsep}{3pt}
\begin{tabular}{lrrrrrr}
\toprule
& \multicolumn{3}{c}{AUROC} & \multicolumn{3}{c}{AP} \\
Class/category & \DEC & \PROB & $\Delta$ & \DEC & \PROB & $\Delta$ \\
\midrule
\multicolumn{7}{l}{\emph{UCF-Crime}} \\
Abuse & 65.13 & 94.73 & +29.60 & 0.14 & 0.33 & +0.18 \\
Arrest & 85.02 & 97.48 & +12.46 & 17.45 & 47.33 & +29.89 \\
Arson & 67.87 & 72.92 & +5.05 & 9.27 & 26.83 & +17.56 \\
Assault & 89.58 & 99.33 & +9.75 & 34.15 & 82.19 & +48.04 \\
Burglary & 67.08 & 89.16 & +22.08 & 10.32 & 32.46 & +22.14 \\
Explosion & 85.14 & 95.84 & +10.70 & 15.27 & 48.30 & +33.04 \\
Fighting & 87.95 & 98.52 & +10.57 & 17.81 & 57.23 & +39.42 \\
RoadAccidents & 86.87 & 96.47 & +9.60 & 9.55 & 29.06 & +19.51 \\
Robbery & 78.37 & 96.32 & +17.94 & 10.89 & 57.42 & +46.53 \\
Shooting & 77.93 & 94.50 & +16.57 & 10.40 & 23.20 & +12.79 \\
Shoplifting & 56.15 & 83.29 & +27.14 & 2.25 & 4.81 & +2.56 \\
Stealing & 76.16 & 96.88 & +20.72 & 11.05 & 45.03 & +33.98 \\
Vandalism & 78.68 & 98.16 & +19.48 & 5.58 & 21.03 & +15.45 \\
\midrule
\multicolumn{7}{l}{\emph{XD-Violence}} \\
Abuse & 89.90 & 99.02 & +9.12 & 10.67 & 49.28 & +38.61 \\
Car accident & 76.65 & 91.68 & +15.04 & 19.37 & 38.01 & +18.64 \\
Explosion & 89.81 & 97.51 & +7.70 & 31.10 & 54.84 & +23.74 \\
Fighting & 89.25 & 96.97 & +7.73 & 45.62 & 66.35 & +20.74 \\
Riot & 83.91 & 96.99 & +13.08 & 60.50 & 87.44 & +26.94 \\
Shooting & 89.99 & 97.11 & +7.12 & 27.30 & 44.35 & +17.05 \\
\bottomrule
\end{tabular}
\end{table}

\subsection{Model sizes}
\label{sec:supp_size}
\Cref{tab:model_size} extends the Yes/No comparison from 2B to 38B parameters while retaining the domain-specific prompts and visual protocol.
Every paired endpoint favors \PROB on both metrics.
The 2B video runs and the 7--8B and larger image controls use RTX PRO 6000 Blackwell hardware.
The 2B/3B image endpoints use RTX 6000 Ada hardware.

\begin{table}[tbp]
\centering
\caption{Yes/No performance across model sizes (\%). 
MVTec AD uses category-macro image metrics. 
Video metrics use frame-level scores.
}
\label{tab:model_size}
\small
\setlength{\tabcolsep}{3pt}
\begin{tabular}{llrrrr}
\toprule
Benchmark & Model & \multicolumn{2}{c}{AUROC} & \multicolumn{2}{c}{AP} \\
& & \DEC & \PROB & \DEC & \PROB \\
\midrule
UCF-Crime & Qwen3-VL-2B & 59.01 & 87.54 & 18.40 & 41.38 \\
UCF-Crime & Qwen3-VL-8B & 76.28 & 85.26 & 19.91 & 40.87 \\
UCF-Crime & Qwen3-VL-32B & 76.22 & 86.84 & 23.48 & 43.63 \\
UCF-Crime & InternVL3.5-2B & 61.23 & 86.87 & 19.63 & 41.91 \\
UCF-Crime & InternVL3.5-8B & 75.97 & 85.93 & 22.70 & 39.39 \\
UCF-Crime & InternVL3.5-38B & 76.60 & 87.47 & 22.43 & 43.87 \\
XD-Violence & Qwen3-VL-2B & 70.09 & 88.92 & 43.56 & 62.60 \\
XD-Violence & Qwen3-VL-8B & 84.64 & 91.40 & 50.66 & 70.25 \\
XD-Violence & InternVL3.5-2B & 77.94 & 91.85 & 53.83 & 73.22 \\
XD-Violence & InternVL3.5-8B & 85.31 & 91.67 & 55.32 & 73.31 \\
MVTec AD & Qwen3-VL-2B & 53.92 & 87.93 & 74.43 & 92.91 \\
MVTec AD & Qwen3-VL-8B & 75.05 & 91.51 & 85.56 & 95.67 \\
MVTec AD & Qwen3-VL-32B & 81.68 & 92.98 & 88.77 & 96.92 \\
MVTec AD & Qwen2.5-VL-3B & 68.09 & 83.81 & 81.45 & 91.26 \\
MVTec AD & Qwen2.5-VL-7B & 59.26 & 90.82 & 77.44 & 95.56 \\
MVTec AD & InternVL3.5-8B & 69.47 & 84.18 & 82.16 & 92.11 \\
MVTec AD & InternVL3.5-38B & 73.15 & 88.65 & 84.62 & 94.51 \\
\bottomrule
\end{tabular}
\end{table}

\subsection{Low-false-positive-rate performance}
\label{sec:supp_operating}
\label{sec:supp_fixedfpr}
We report the highest TPR attainable subject to an FPR budget:
\begin{equation}
\operatorname{TPR}^{*}(\alpha)
 = \max_{t:\operatorname{FPR}(t)\leq\alpha}\operatorname{TPR}(t).
\label{eq:supp_fixedfpr}
\end{equation}
We include the reject-all prediction, process equal scores together, and do not interpolate.
Image thresholds are selected within category before macro-averaging.
Video thresholds use concatenated frame-level scores.
These thresholds are selected using test labels to characterize the attainable operating points.
Mean TPR improves on every benchmark at every budget.
At the individual-configuration level, \PROB is higher in all 28 UCF-Crime comparisons at each budget and in 27, 28, and 28 XD-Violence comparisons at 1\%, 5\%, and 10\%, respectively.

\begin{table}[tbp]
\centering
\caption{Attainable TPR (\%) at fixed FPR budgets, averaged over 28 model--scale configurations. Image results are category-macro and video results are frame-level.}
\label{tab:image_operating_point}
\label{tab:video_operating_point}
\small
\begin{tabular}{llrrr}
\toprule
Benchmark & Rule & FPR $\leq 1\%$ & FPR $\leq 5\%$ & FPR $\leq 10\%$ \\
\midrule
MVTec AD & \DEC & 34.91 & 40.55 & 42.84 \\
 & \PROB & 61.36 & 70.72 & 76.10 \\
VisA & \DEC & 19.68 & 23.59 & 24.85 \\
 & \PROB & 42.04 & 56.63 & 63.36 \\
UCF-Crime & \DEC & 1.00 & 17.37 & 38.94 \\
 & \PROB & 16.46 & 43.59 & 57.86 \\
XD-Violence & \DEC & 0.33 & 7.59 & 24.46 \\
 & \PROB & 10.56 & 39.61 & 66.54 \\
\bottomrule
\end{tabular}
\end{table}

\section{Rank Compression}
\label{app:mechanism}

\subsection{Tie-breaking recovery}
\label{sec:supp_decomp}
We rank by $H=(s_{\mathrm{DEC}},s_{\mathrm{PROB}})$ lexicographically in descending order.
Thus \PROB orders inputs only within a \DEC tie.
For metric $M$, define
\[
\Delta=M(\PROB)-M(\DEC),\qquad
T=M(H)-M(\DEC),\qquad
R=M(\PROB)-M(H).
\]
Then $\Delta=T+R$.
For a set of configurations $S$, recovery is $100\,\overline{T}_S/\overline{\Delta}_S$.
A value above 100\% means that preserving the order between decoded levels outperforms the full \PROB ranking.
For binary scales, tie-breaking reconstructs the \PROB ranking by construction.
The nonbinary rows measure the empirical contribution of within-answer ordering.

\begin{table}[tbp]
\centering
\caption{Primary-metric recovery (\%). Rows average four models before computing the ratio of gains. All 7 and Nonbinary 5 average 28 and 20 configurations. Columns use I-AUROC, I-AUROC, AUROC, and AP, respectively.}
\label{tab:recovery_scales}
\small
\setlength{\tabcolsep}{3pt}
\begin{tabular}{lrrrr}
\toprule
Scale & MVTec AD & VisA & UCF-Crime & XD-Violence \\
\midrule
Yes/No & 100.00 & 100.00 & 100.00 & 100.00 \\
0--1 integer & 100.00 & 100.00 & 100.00 & 100.00 \\
0--5 integer & 98.91 & 99.95 & 96.83 & 97.78 \\
0--9 integer & 98.69 & 100.67 & 95.35 & 90.26 \\
0--1 decimal & 99.13 & 100.13 & 97.84 & 93.05 \\
0--5 decimal & 100.19 & 100.93 & 85.16 & 93.30 \\
0--9 decimal & 99.58 & 102.43 & 73.26 & 75.46 \\
All 7 scales & 99.57 & 100.48 & 95.44 & 95.19 \\
Nonbinary 5 & 99.27 & 100.77 & 91.50 & 90.70 \\
\bottomrule
\end{tabular}
\end{table}

\begin{table}[tbp]
\centering
\caption{Primary-metric recovery (\%) by model. All 7 and Nonbinary 5 are ratios of mean gains across their answer scales. The last column reports the finest scale.}
\label{tab:recovery_models}
\small
\setlength{\tabcolsep}{3pt}
\begin{tabular}{llrrr}
\toprule
Benchmark & Model & All 7 & Nonbinary 5 & 0--9 decimal \\
\midrule
MVTec AD & Qwen3-VL-8B & 100.09 & 100.15 & 100.06 \\
MVTec AD & Qwen2.5-VL-7B & 99.72 & 99.41 & 99.62 \\
MVTec AD & InternVL3.5-8B & 98.67 & 97.73 & 98.95 \\
MVTec AD & MiniCPM-V-4.5-8B & 99.84 & 99.77 & 99.68 \\
VisA & Qwen3-VL-8B & 100.04 & 100.07 & 100.12 \\
VisA & Qwen2.5-VL-7B & 99.94 & 99.89 & 100.22 \\
VisA & InternVL3.5-8B & 101.82 & 103.06 & 112.76 \\
VisA & MiniCPM-V-4.5-8B & 100.31 & 100.43 & 100.57 \\
UCF-Crime & Qwen3-VL-8B & 99.51 & 98.85 & 96.85 \\
UCF-Crime & Qwen2.5-VL-7B & 93.56 & 89.18 & 68.27 \\
UCF-Crime & InternVL3.5-8B & 92.31 & 79.37 & 67.92 \\
UCF-Crime & MiniCPM-V-4.5-8B & 99.30 & 98.90 & 93.57 \\
XD-Violence & Qwen3-VL-8B & 99.28 & 98.36 & 93.99 \\
XD-Violence & Qwen2.5-VL-7B & 85.17 & 73.01 & 37.90 \\
XD-Violence & InternVL3.5-8B & 99.47 & 99.00 & 103.25 \\
XD-Violence & MiniCPM-V-4.5-8B & 96.57 & 93.74 & 81.32 \\
\bottomrule
\end{tabular}
\end{table}

Recovery varies across configurations.
At the finest video scale, the four-model values are 73.26\% on UCF-Crime and 75.46\% on XD-Violence.
The lowest model-level value in \Cref{tab:recovery_models} is 37.90\% for Qwen2.5-VL-7B on XD-Violence at this scale.

\subsection{Ordering within each decoded answer}
\Cref{tab:within_answer} evaluates \PROB among inputs assigned the same Yes/No answer.
All 16 model--benchmark ``No'' results are above chance.
Image averages include categories containing both labels in the corresponding answer group, with their counts in parentheses.
The ``Yes'' results include a below-chance value of 46.90 on XD-Violence for Qwen2.5-VL-7B.

\begin{table}[tbp]
\centering
\caption{Within-answer AUROC under Yes/No (\%). Image results are category-macro, with the number of eligible categories in parentheses. Video results are frame-level.}
\label{tab:within_answer}
\small
\setlength{\tabcolsep}{3pt}
\begin{tabular}{lcccc}
\toprule
Model & \multicolumn{2}{c}{MVTec AD} & \multicolumn{2}{c}{VisA} \\
& No & Yes & No & Yes \\
\midrule
Qwen3-VL-8B      & 83.99 (15) & 85.24 (3) & 85.47 (12) & 62.49 (3) \\
Qwen2.5-VL-7B    & 88.59 (15) & 90.00 (1) & 80.61 (12) & 50.00 (1) \\
InternVL3.5-8B   & 76.80 (15) & 72.26 (4) & 77.29 (12) & 83.42 (2) \\
MiniCPM-V-4.5-8B & 82.81 (15) & 98.91 (1) & 73.89 (12) & 53.87 (1) \\
\midrule
Model & \multicolumn{2}{c}{UCF-Crime} & \multicolumn{2}{c}{XD-Violence} \\
& No & Yes & No & Yes \\
Qwen3-VL-8B      & 76.23 & 71.36 & 82.43 & 72.77 \\
Qwen2.5-VL-7B    & 83.13 & 60.84 & 89.38 & 46.90 \\
InternVL3.5-8B   & 76.88 & 69.51 & 84.22 & 69.38 \\
MiniCPM-V-4.5-8B & 71.33 & 70.84 & 83.49 & 70.00 \\
\bottomrule
\end{tabular}
\end{table}

\subsection{Oracle binary thresholds}
\label{sec:supp_threshold}
We evaluate every binary partition induced by the Yes/No \PROB scores, including accept-all and reject-all, and choose the test-set threshold maximizing the primary metric.
Equal scores remain grouped.
Images use one threshold per category to maximize I-AUROC.
Videos use one threshold per model to maximize UCF-Crime AUROC or XD-Violence AP.
\Cref{tab:oracle_threshold} shows that the oracle improves on \DEC but remains below \PROB in all 16 settings.

\begin{table*}[tbp]
\centering
\caption{Oracle binary-threshold diagnostic under Yes/No (\%). Test labels select the threshold maximizing the primary metric. Threshold gain is Oracle binary minus \DEC, and residual gap is \PROB minus Oracle binary, in percentage points. Differences use unrounded values.}
\label{tab:oracle_threshold}
\small
\setlength{\tabcolsep}{4pt}
\resizebox{\textwidth}{!}{%
\begin{tabular}{llrrrrr}
\toprule
Benchmark & Model & \DEC & Oracle binary & \PROB & Threshold gain & Residual gap \\
\midrule
\multirow{4}{*}{MVTec AD} & Qwen3-VL-8B & 75.05 & 88.07 & 91.51 & 13.02 & 3.44 \\
 & Qwen2.5-VL-7B & 59.26 & 86.85 & 90.82 & 27.58 & 3.97 \\
 & InternVL3.5-8B & 69.47 & 82.39 & 84.18 & 12.91 & 1.79 \\
 & MiniCPM-V-4.5-8B & 74.73 & 87.89 & 89.25 & 13.17 & 1.36 \\
\midrule
\multirow{4}{*}{VisA} & Qwen3-VL-8B & 62.00 & 82.96 & 87.23 & 20.96 & 4.27 \\
 & Qwen2.5-VL-7B & 53.96 & 76.41 & 81.27 & 22.46 & 4.85 \\
 & InternVL3.5-8B & 58.42 & 76.57 & 79.47 & 18.15 & 2.91 \\
 & MiniCPM-V-4.5-8B & 59.38 & 75.86 & 76.27 & 16.48 & 0.41 \\
\midrule
\multirow{4}{*}{UCF-Crime} & Qwen3-VL-8B & 76.28 & 77.66 & 85.26 & 1.38 & 7.60 \\
 & Qwen2.5-VL-7B & 60.47 & 78.55 & 85.64 & 18.08 & 7.09 \\
 & InternVL3.5-8B & 75.97 & 78.89 & 85.93 & 2.92 & 7.03 \\
 & MiniCPM-V-4.5-8B & 76.07 & 77.15 & 84.45 & 1.08 & 7.30 \\
\midrule
\multirow{4}{*}{XD-Violence} & Qwen3-VL-8B & 50.66 & 55.65 & 70.25 & 4.99 & 14.60 \\
 & Qwen2.5-VL-7B & 36.81 & 51.20 & 59.19 & 14.39 & 7.99 \\
 & InternVL3.5-8B & 55.32 & 56.61 & 73.31 & 1.29 & 16.70 \\
 & MiniCPM-V-4.5-8B & 54.97 & 55.86 & 73.17 & 0.89 & 17.31 \\
\bottomrule
\end{tabular}%
}
\end{table*}

\subsection{Distinct scores and numerical precision}
\label{app:scale}
Under the 0--9 decimal scale, \DEC produces only 3.2--5.8 distinct scores per image category and 9--18 over a video benchmark.
\Cref{tab:prob_unique_coverage} reports the corresponding \PROB coverage across the decimal scales.
The remaining video duplicates arise from repeated inputs near video boundaries and numerically identical probabilities.

\begin{table}[tbp]
\centering
\caption{Distinct \PROB scores across three decimal scales and four models. Image counts are means per category. Video counts are prediction segments. Ranges span the 12 configurations.}
\label{tab:prob_unique_coverage}
\setlength{\tabcolsep}{2.5pt}
\begin{tabular}{lrrr}
\toprule
Benchmark & Units & Unique \PROB & Unique (\%) \\
\midrule
MVTec AD   & 115.0   & 115.0 & 100.000 \\
VisA       & 180.2   & 180.2 & 100.000 \\
UCF-Crime  & 69,634  & 69,538--69,542 & 99.862--99.868 \\
XD-Violence& 146,449 & 146,334--146,421 & 99.921--99.981 \\
\bottomrule
\end{tabular}
\end{table}

\paragraph{Precision comparison.}
\label{sec:supp_precision}
For Qwen3-VL-8B with Yes/No on UCF-Crime, switching from bfloat16 to float32 increases distinct \PROB scores from 137 to 65,885, while AUROC changes from 85.26 to 85.27.
The segment-score Spearman correlation is 0.9985.
On MVTec AD, the mean count per category rises from 53.1 to 115.0, I-AUROC changes from 91.51 to 91.78, and I-AP from 95.67 to 95.98.
The runs also differ in batch size.
The comparison demonstrates that a large increase in distinct-score count need not substantially improve ranking.

\section{Additional Scoring Experiments}
\label{app:controls}

\subsection{Sampled answers}
\label{sec:supp_sampling}
Single-answer video simulations use temperatures 0.6 and 1.0, nucleus top-$p$ values 0.8 and 0.95 at $T=1$, and majority votes over 5 and 16 samples at $T=1$.
A tied 16-sample vote returns No.
All draw from saved answer-set-normalized Yes/No probabilities.
Across these strategies, four-model mean performance ranges from 70.85 to 72.13 UCF-Crime AUROC and from 46.03 to 49.09 XD-Violence AP, below the corresponding \PROB values of 85.32 and 68.98.

For sample-mean scoring, we use $T=1$ and $k\in\{1,2,4,8,16,32,64,128,256\}$ and score each input by its fraction of Yes answers.
Metrics are computed for each model--seed run before averaging over three seeds and four VLMs.
Image seeds are 20260903, 20260904, and 20260905.
Models share uniform random streams within each dataset and seed, and smaller budgets reuse prefixes of 256 draws.
Video seeds are 0, 1, and 2, with binomial draws for each segment and sample count before frame expansion.
\Cref{tab:image_sampling} reports the image results, and \Cref{fig:sampled_answers} shows all four benchmarks.

\begin{table}[tbp]
\centering
\caption{Sample-mean image performance under Yes/No (\%). Metrics are evaluated per model and seed before averaging. \DEC and \PROB are the four-model means without sampling.}
\label{tab:image_sampling}
\small
\begin{tabular}{lrrrr}
\toprule
& \multicolumn{2}{c}{MVTec AD} & \multicolumn{2}{c}{VisA} \\
\cmidrule(lr){2-3}\cmidrule(lr){4-5}
Scoring rule / samples & I-AUROC & I-AP & I-AUROC & I-AP \\
\midrule
Greedy \DEC & 69.63 & 82.67 & 58.44 & 63.27 \\
$k=1$       & 68.47 & 81.84 & 59.28 & 63.61 \\
$k=2$       & 72.64 & 84.25 & 62.35 & 66.35 \\
$k=4$       & 76.12 & 86.50 & 65.09 & 69.15 \\
$k=8$       & 78.98 & 88.40 & 67.95 & 71.87 \\
$k=16$      & 81.18 & 89.90 & 69.95 & 74.02 \\
$k=32$      & 83.17 & 91.39 & 71.79 & 75.99 \\
$k=64$      & 84.64 & 92.42 & 73.91 & 78.10 \\
$k=128$     & 85.72 & 93.12 & 75.53 & 79.76 \\
$k=256$     & 86.68 & 93.72 & 76.69 & 80.83 \\
Direct \PROB & 88.94 & 94.71 & 81.06 & 84.35 \\
\bottomrule
\end{tabular}
\end{table}

\subsection{Temporal smoothing}
\label{sec:supp_sigma}
\label{app:temporal_smoothing}
We apply matched Gaussian smoothing at $\sigma\in\{1,\ldots,9\}$ to each video's segment scores before frame expansion.
The kernel is truncated at $4\sigma$ with reflected boundaries.
Mean primary-metric gaps remain positive at every width (\Cref{tab:smoothing_endpoints}).
Across individual model--scale--width comparisons, \PROB is higher in 207 of 252 UCF-Crime AUROC comparisons and 230 of 252 XD-Violence AP comparisons.
Distinct scores are counted by exact float64 equality without rounding.

\begin{table}[tbp]
\centering
\caption{Matched smoothing: primary-metric gap in percentage points and distinct segment-score counts, averaged over 28 configurations. Rows show representative widths and the unsmoothed baseline.}
\label{tab:smoothing_endpoints}
\small
\begin{tabular}{lrrrr}
\toprule
Benchmark & $\sigma$ & \PROB--\DEC gap & Unique \DEC & Unique \PROB \\
\midrule
UCF-Crime & 0 & 7.664 & 7.0 & 46,424.9 \\
& 1 & 4.613 & 4,989.0 & 69,479.1 \\
& 4 & 2.947 & 22,979.5 & 69,555.4 \\
& 8 & 2.350 & 30,549.3 & 69,561.2 \\
& 9 & 2.284 & 31,744.9 & 69,561.9 \\
\midrule
XD-Violence & 0 & 10.972 & 7.7 & 99,740.4 \\
& 1 & 5.742 & 10,745.4 & 145,665.5 \\
& 4 & 3.862 & 57,905.2 & 146,256.4 \\
& 8 & 3.131 & 75,080.5 & 146,343.2 \\
& 9 & 3.008 & 77,582.8 & 146,356.2 \\
\bottomrule
\end{tabular}
\end{table}

\subsection{Prompt wording and answer polarity}
\label{sec:supp_prompt_sens}
\label{sec:supp_polarity}
We evaluate three paraphrases and a reversed-normality question using Qwen3-VL-8B and InternVL3.5-8B on MVTec AD and UCF-Crime.
Paraphrases preserve the Yes/No answer set, value map, visual protocol, and batch size.
The reversed question assigns $v(\text{No})=1$ and $v(\text{Yes})=0$.
\PROB exceeds \DEC on both metrics in every row of \Cref{tab:prompt_controls}.

\paragraph{Video pv1.}
\begin{quote}\small\ttfamily\raggedright
You are reviewing ordered frames from a surveillance video clip.\\
Is anything unusual or dangerous happening in the clip?\\
Answer with exactly one word: Yes or No.
\end{quote}

\paragraph{Video pv2.}
\begin{quote}\small\ttfamily\raggedright
You are a security analyst monitoring surveillance footage.\\
Do these ordered frames show an anomalous, suspicious, violent, criminal, or dangerous event?\\
Answer with exactly one word: Yes or No.
\end{quote}

\paragraph{Video pv3.}
\begin{quote}\small\ttfamily\raggedright
These are ordered frames from a surveillance video clip.\\
Does the clip contain an anomaly, such as violence, crime, or a dangerous situation?\\
Answer with exactly one word: Yes or No.
\end{quote}

\paragraph{Image pv1.}
\begin{quote}\small\ttfamily\raggedright
You are inspecting an image of a <category>.\\
Is the <category> unusual or defective in any way?\\
Answer with exactly one word: Yes or No.
\end{quote}

\paragraph{Image pv2.}
\begin{quote}\small\ttfamily\raggedright
You are a quality-control inspector examining an image of a <category>.\\
Does the object or material show an anomaly, damage, contamination, malformation, misplacement, or a missing part?\\
Answer with exactly one word: Yes or No.
\end{quote}

\paragraph{Image pv3.}
\begin{quote}\small\ttfamily\raggedright
This image shows a <category>.\\
Does it contain a defect, such as damage, contamination, an incorrect shape, a misplaced component, or a missing part?\\
Answer with exactly one word: Yes or No.
\end{quote}

\paragraph{Image reversed-polarity prompt.}
\begin{quote}\small\ttfamily\raggedright
You are inspecting an image of a <category>.\\
Is the object or material normal and defect-free, with no damage, contamination, malformation, misplacement, or missing part?\\
Answer with exactly one word: Yes or No.
\end{quote}
For this control, No maps to 1 and Yes maps to 0.

\paragraph{Video reversed-polarity prompt.}
\begin{quote}\small\ttfamily\raggedright
You are reviewing ordered frames from a surveillance video clip.\\
Is the clip normal, ordinary, and free of anything suspicious, violent, criminal, or dangerous?\\
Answer with exactly one word: Yes or No.
\end{quote}
The same reversed value map applies.

\begin{table}[tbp]
\centering
\caption{Prompt and polarity controls (\%). Base and pv1--pv3 assign anomaly to Yes. Reversed assigns anomaly to No. Image metrics are category-macro and video metrics are unsmoothed frame-level.}
\label{tab:prompt_controls}
\small
\setlength{\tabcolsep}{3pt}
\begin{tabular}{lllrrrr}
\toprule
Benchmark & Model & Control & \multicolumn{2}{c}{AUROC} & \multicolumn{2}{c}{AP} \\
& & & \DEC & \PROB & \DEC & \PROB \\
\midrule
MVTec AD & Qwen3-VL-8B & base & 75.05 & 91.51 & 85.56 & 95.67 \\
MVTec AD & Qwen3-VL-8B & pv1 & 77.13 & 91.34 & 86.70 & 95.57 \\
MVTec AD & Qwen3-VL-8B & pv2 & 79.63 & 92.17 & 87.61 & 95.92 \\
MVTec AD & Qwen3-VL-8B & pv3 & 79.78 & 92.31 & 87.67 & 95.84 \\
MVTec AD & InternVL3.5-8B & base & 69.47 & 84.18 & 82.16 & 92.11 \\
MVTec AD & InternVL3.5-8B & pv1 & 68.48 & 85.30 & 81.68 & 92.87 \\
MVTec AD & InternVL3.5-8B & pv2 & 71.89 & 83.92 & 83.14 & 91.81 \\
MVTec AD & InternVL3.5-8B & pv3 & 72.44 & 83.80 & 83.45 & 92.07 \\
UCF-Crime & Qwen3-VL-8B & base & 76.28 & 85.26 & 19.91 & 40.87 \\
UCF-Crime & Qwen3-VL-8B & pv1 & 77.40 & 85.95 & 19.82 & 40.83 \\
UCF-Crime & Qwen3-VL-8B & pv2 & 74.94 & 86.03 & 22.72 & 43.43 \\
UCF-Crime & Qwen3-VL-8B & pv3 & 75.96 & 85.30 & 20.85 & 41.40 \\
UCF-Crime & InternVL3.5-8B & base & 75.97 & 85.93 & 22.70 & 39.39 \\
UCF-Crime & InternVL3.5-8B & pv1 & 75.91 & 84.71 & 22.12 & 39.60 \\
UCF-Crime & InternVL3.5-8B & pv2 & 73.98 & 84.60 & 22.41 & 38.51 \\
UCF-Crime & InternVL3.5-8B & pv3 & 74.55 & 85.12 & 21.89 & 38.80 \\
MVTec AD & Qwen3-VL-8B & Reversed & 72.08 & 90.63 & 82.90 & 95.28 \\
MVTec AD & InternVL3.5-8B & Reversed & 71.98 & 82.15 & 83.36 & 90.39 \\
UCF-Crime & Qwen3-VL-8B & Reversed & 77.77 & 85.06 & 18.31 & 39.02 \\
UCF-Crime & InternVL3.5-8B & Reversed & 76.66 & 85.84 & 21.74 & 35.63 \\
\bottomrule
\end{tabular}
\end{table}

\subsection{Answers accompanied by descriptions}
\label{sec:supp_explain}
We evaluate Qwen3-VL-8B and MiniCPM-V-4.5-8B on both binary scales with a factual description before or after the answer.
Descriptions use greedy decoding (\texttt{do\_sample=False}, \texttt{num\_beams=1}) with at most 64 tokens.
Score-first conditions the description on the argmax answer.
Explanation-first conditions both scoring rules on the same generated description.
\Cref{tab:joint_explain} reports the primary-metric comparisons.

\paragraph{Image joint-output prompt.}
The following common stem is followed by one score instruction and one output-order instruction, separated by newlines.
\begin{quote}\small\ttfamily\raggedright
You are inspecting an image of a <category>.\\
Judge whether the object or material is anomalous or defective, for example damaged, contaminated, malformed, misplaced, or missing a part.\\
Also describe only what is visibly present in one concise factual sentence.\\
Do not speculate about causes or anything that is not visible.
\end{quote}
For Yes/No, the score instruction is:
\begin{quote}\small\ttfamily\raggedright
The score must be exactly Yes or No: No means normal and Yes means anomalous or defective.
\end{quote}
For 0--1 integer, it is:
\begin{quote}\small\ttfamily\raggedright
The score must be exactly 0 or 1: 0 means normal and 1 means anomalous or defective.
\end{quote}
Score-first uses:
\begin{quote}\small\ttfamily\raggedright
Return exactly two lines in this order:\\
Score: <Yes or No>\\
Description: <one concise factual sentence>
\end{quote}
Explanation-first uses:
\begin{quote}\small\ttfamily\raggedright
Return exactly two lines in this order:\\
Description: <one concise factual sentence>\\
Score: <Yes or No>
\end{quote}
For the integer scale, the literal placeholder \texttt{<Yes or No>} is replaced by \texttt{<0 or 1>}. These placeholders appear literally in the prompt. Only \texttt{<category>} is substituted with a category name.
The prompt version is \texttt{industrial-image-joint-v1} with the output order recorded in each run.

\paragraph{Video joint-output prompt.}

For video, replace the image common stem above with:
\begin{quote}\small\ttfamily\raggedright
You are reviewing ordered frames from a surveillance video clip.\\
Judge whether the clip is anomalous, suspicious, violent, criminal, or dangerous.\\
Also describe only what is visibly happening in one concise factual sentence.\\
Do not speculate about intent or events that are not visible.
\end{quote}
The video Yes/No instruction is \texttt{The score must be exactly Yes or No.}
The video integer instruction is \texttt{The score must be exactly 0 or 1: 0 means normal and 1 means anomalous or dangerous.}
Append the same output-order instructions shown above, with newline separators.

\begin{table}[tbp]
\centering
\caption{Primary-metric performance with descriptions (\%). S-first emits the score first, and E-first emits the description first. MiniCPM-V denotes MiniCPM-V-4.5-8B.}
\label{tab:joint_explain}
\small
\setlength{\tabcolsep}{3pt}
\begin{tabular}{lllrrrrrr}
\toprule
Benchmark & Model & Scale & \multicolumn{2}{c}{Base} & \multicolumn{2}{c}{S-first} & \multicolumn{2}{c}{E-first} \\
& & & \DEC & \PROB & \DEC & \PROB & \DEC & \PROB \\
\midrule
UCF-Crime & Qwen3-VL-8B      & Yes/No & 76.28 & 85.26 & 76.08 & 85.07 & 75.12 & 84.46 \\
UCF-Crime & Qwen3-VL-8B      & 0--1 int & 75.23 & 85.49 & 76.56 & 84.96 & 75.18 & 84.37 \\
UCF-Crime & MiniCPM-V-4.5-8B & Yes/No & 76.07 & 84.45 & 74.64 & 82.34 & 75.72 & 83.38 \\
UCF-Crime & MiniCPM-V-4.5-8B & 0--1 int & 76.60 & 84.10 & 76.07 & 83.39 & 74.20 & 81.94 \\
XD-Violence & Qwen3-VL-8B      & Yes/No & 50.66 & 70.25 & 51.11 & 71.50 & 48.33 & 65.03 \\
XD-Violence & Qwen3-VL-8B      & 0--1 int & 43.35 & 69.42 & 47.67 & 64.44 & 47.56 & 58.86 \\
XD-Violence & MiniCPM-V-4.5-8B & Yes/No & 54.97 & 73.17 & 53.96 & 73.18 & 53.26 & 70.92 \\
XD-Violence & MiniCPM-V-4.5-8B & 0--1 int & 51.01 & 71.84 & 48.81 & 71.90 & 51.30 & 67.91 \\
MVTec AD & Qwen3-VL-8B & Yes/No   & 75.05 & 91.51 & 65.62 & 92.57 & 67.60 & 88.49 \\
MVTec AD & Qwen3-VL-8B & 0--1 int & 67.96 & 91.46 & 75.52 & 92.07 & 73.49 & 87.85 \\
MVTec AD & MiniCPM-V    & Yes/No   & 74.73 & 89.25 & 75.48 & 93.12 & 72.21 & 91.89 \\
MVTec AD & MiniCPM-V    & 0--1 int & 72.20 & 93.18 & 79.90 & 93.39 & 79.69 & 90.28 \\
VisA & Qwen3-VL-8B & Yes/No   & 62.00 & 87.23 & 55.96 & 87.27 & 56.92 & 80.19 \\
VisA & Qwen3-VL-8B & 0--1 int & 58.04 & 87.79 & 61.92 & 86.75 & 62.42 & 81.64 \\
VisA & MiniCPM-V    & Yes/No   & 59.38 & 76.27 & 61.29 & 84.04 & 58.92 & 83.79 \\
VisA & MiniCPM-V    & 0--1 int & 58.08 & 85.50 & 72.81 & 86.27 & 67.59 & 83.45 \\
\bottomrule
\end{tabular}
\end{table}

\PROB remains higher in every paired table comparison.
Output order changes absolute performance.
Across the eight unsmoothed video configurations, explanation-first reduces \PROB by 3.39 primary-metric points on average, including a 10.57-point AP decrease.
The paired scoring comparison therefore remains distinct from the effect of generating a description.

\section{Evaluation with Ties}
\label{app:ap}

\subsection{Benchmark evaluation code}
\label{sec:supp_official_metrics}
\Cref{tab:official_tie_conventions} identifies the exact releases and metric entry points inspected for the main-text comparison.

\begin{table}[tbp]
\centering
\caption{Metric conventions in the inspected releases. Absent means that the metric is not reported in the specified path.}
\label{tab:official_tie_conventions}
\small
\setlength{\tabcolsep}{3pt}
\begin{tabular}{>{\raggedright\arraybackslash}p{0.13\linewidth}>{\raggedright\arraybackslash}p{0.20\linewidth}>{\raggedright\arraybackslash}p{0.20\linewidth}>{\raggedright\arraybackslash}p{0.37\linewidth}}
\toprule
Benchmark & ROC & PR & Release / metric entry point \\
\midrule
MVTec AD & Grouped, half-credit & Absent at image level & Evaluation archive v1.0. \nolinkurl{roc_curve_util.py}, \nolinkurl{compute_classification_roc}. \\
VisA & Grouped, half-credit & Grouped, trapezoidal & \href{https://github.com/amazon-research/spot-diff/tree/e1429125dbaa3406faaf42626ad17964ed47718d}{spot-diff, \texttt{e1429125dbaa}}. \nolinkurl{utils/metrics.py}, \nolinkurl{compute_classification_metrics}. \\
UCF-Crime & Item-wise, order-dependent & Absent & \href{https://github.com/WaqasSultani/AnomalyDetectionCVPR2018/tree/5e3ee7a991ce8a42cd30eb7293734f56536f3df1}{author release, \texttt{5e3ee7a991ce}}. \nolinkurl{Evaluate_Anomaly_Detector.m}, active ROC block. \\
XD-Violence & Absent & Grouped, trapezoidal & \href{https://github.com/Roc-Ng/XDVioDet/tree/f846d8cfc454943ec165bd2489b71e7dc0a1064d}{XDVioDet, \texttt{f846d8cfc454}}. \nolinkurl{test.py}, offline/online statements. \\
\bottomrule
\end{tabular}
\end{table}

Checks of the MVTec AD and VisA functions and XD-Violence metric statements matched the grouped calculations in 384 tests per release, using 16 score assignments and 24 input permutations.
The UCF-Crime check used a literal NumPy translation of the active MATLAB ROC block, not native MATLAB execution.
The audit environment was Python 3.11.15, NumPy 2.4.4, scikit-learn 1.9.0, and SciPy 1.17.1.

\subsection{Exact ROC tie-order sensitivity}
\label{sec:supp_auc_ties}
Let $p_j$ and $n_j$ count anomalous and normal examples in tied-score group $j$, and let $P=\sum_j p_j>0$ and $N=\sum_j n_j>0$.
The fraction of tied anomalous--normal pairs is
\begin{equation}
\tau=\frac{\sum_j p_jn_j}{PN}.
\label{eq:supp_tied_pair_fraction}
\end{equation}
Grouping ties gives the half-credit AUROC in \Cref{eq:auc}.
An item-wise ROC traversal instead assigns an order to each tied pair.
Putting anomalous examples before all normal examples within each group gives each such pair full credit, while the reverse order gives zero.
All untied pairs retain their ordering, so the exact extrema are
\begin{equation}
\mathrm{AUROC}^{-}=\mathrm{AUROC}-\frac{\tau}{2},
\qquad
\mathrm{AUROC}^{+}=\mathrm{AUROC}+\frac{\tau}{2}.
\label{eq:supp_auc_order}
\end{equation}
The maximum item-wise ROC-area span is therefore $\tau$.
These label-aware extrema measure worst-case sensitivity to an unspecified tie order, not uncertainty in the grouped AUROC.
\Cref{tab:ap_tie_sensitivity} reports the tied-pair fractions on the saved scores.
Image fractions are computed within category before macro-averaging.
Video fractions use frame-level scores, so segment-to-frame repetition can create mixed-label ties at anomaly boundaries.

\begin{table}[tbp]
\centering
\caption{Mean fraction of tied anomalous--normal pairs (\%) across four models and seven scales. Numerically, each value is also the maximum item-wise ROC-area span in percentage points.}
\label{tab:ap_tie_sensitivity}
\small
\begin{tabular}{lrr}
\toprule
Benchmark & \DEC & \PROB \\
\midrule
MVTec AD & 49.97 & 0.47 \\
VisA & 71.23 & 1.10 \\
UCF-Crime & 31.32 & 0.23 \\
XD-Violence & 19.29 & 0.20 \\
\bottomrule
\end{tabular}
\end{table}

\subsection{PR-area comparison on fixed scores}
\label{sec:supp_estimator}
\label{sec:supp_tie_policy}
Using the grouped precision--recall points and endpoint in \Cref{eq:pr_area_comparison},
\[
\mathrm{PR\mbox{-}AUC}_{\mathrm{trapezoids}}-\mathrm{AP}
=\frac{1}{2}\sum_{j=1}^{J}(r_j-r_{j-1})(q_{j-1}-q_j).
\]
The difference can have either sign and need not vanish in the absence of ties.
We retain all threshold points in score order, including equal-recall points, without applying a precision envelope.
\Cref{tab:ap_estimator} reports both integration offsets and the resulting \PROB--\DEC gaps for all benchmarks and scales.

\begin{table}[tbp]
\centering
\caption{PR-area calculations on identical scores. Each offset is trapezoidal PR-AUC minus AP for the indicated rule. Each gap is \PROB minus \DEC. Entries are percentage-point differences averaged over four models before rounding. Images use category-macro metrics, and videos use unsmoothed frame-level metrics.}
\label{tab:ap_estimator}
\label{tab:image_ap_estimator}
\small
\setlength{\tabcolsep}{4pt}
\begin{tabular}{@{}llrrrr@{}}
\toprule
Benchmark & Scale & \shortstack{Offset\\\DEC} & \shortstack{Offset\\\PROB}
& \shortstack{Grouped\\AP gap} & \shortstack{Trapezoidal\\PR-AUC gap} \\
\midrule
MVTec AD & Yes/No       & $+8.36$  & $+0.25$ & $+12.05$ & $+3.93$ \\
         & 0--1 integer & $+10.71$ & $+0.17$ & $+16.68$ & $+6.14$ \\
         & 0--5 integer & $+5.84$  & $-0.05$ & $+8.69$  & $+2.80$ \\
         & 0--9 integer & $+5.92$  & $-0.06$ & $+8.54$  & $+2.56$ \\
         & 0--1 decimal & $+7.56$  & $-0.06$ & $+12.05$ & $+4.43$ \\
         & 0--5 decimal & $+5.60$  & $-0.06$ & $+7.85$  & $+2.20$ \\
         & 0--9 decimal & $+5.54$  & $-0.06$ & $+7.83$  & $+2.23$ \\
\midrule
VisA & Yes/No       & $+17.20$ & $+0.79$ & $+21.09$ & $+4.67$ \\
     & 0--1 integer & $+18.79$ & $+0.46$ & $+25.55$ & $+7.22$ \\
     & 0--5 integer & $+13.24$ & $-0.10$ & $+17.11$ & $+3.78$ \\
     & 0--9 integer & $+13.11$ & $-0.11$ & $+16.54$ & $+3.33$ \\
     & 0--1 decimal & $+14.50$ & $-0.11$ & $+20.13$ & $+5.52$ \\
     & 0--5 decimal & $+14.33$ & $-0.11$ & $+16.71$ & $+2.26$ \\
     & 0--9 decimal & $+13.37$ & $-0.11$ & $+16.00$ & $+2.53$ \\
\midrule
UCF-Crime & Yes/No       & $+25.99$ & $+0.45$ & $+18.91$ & $-6.63$ \\
          & 0--1 integer & $+26.08$ & $+0.95$ & $+18.11$ & $-7.02$ \\
          & 0--5 integer & $+15.40$ & $-0.01$ & $+11.49$ & $-3.92$ \\
          & 0--9 integer & $+11.20$ & $-0.01$ & $+11.48$ & $+0.27$ \\
          & 0--1 decimal & $+11.78$ & $-0.02$ & $+11.00$ & $-0.79$ \\
          & 0--5 decimal & $+9.46$  & $-0.02$ & $+9.67$  & $+0.20$ \\
          & 0--9 decimal & $+4.40$  & $-0.01$ & $+8.58$  & $+4.16$ \\
\midrule
XD-Violence & Yes/No       & $+20.74$ & $+0.30$ & $+19.54$ & $-0.89$ \\
            & 0--1 integer & $+22.27$ & $+1.25$ & $+17.55$ & $-3.47$ \\
            & 0--5 integer & $+15.09$ & $-0.00$ & $+9.38$  & $-5.71$ \\
            & 0--9 integer & $+9.25$  & $-0.00$ & $+8.23$  & $-1.03$ \\
            & 0--1 decimal & $+8.98$  & $-0.00$ & $+8.34$  & $-0.63$ \\
            & 0--5 decimal & $+8.84$  & $-0.00$ & $+7.12$  & $-1.72$ \\
            & 0--9 decimal & $+5.54$  & $-0.00$ & $+6.63$  & $+1.09$ \\
\bottomrule
\end{tabular}
\end{table}

Non-interpolated AP is a weighted mean of threshold precisions and therefore cannot exceed their maximum over nonempty predictions.
For image macro-averages, the corresponding bound is the mean of the category-specific maximum precisions.
Trapezoidal PR-AUC exceeds this bound in 1 of 28 MVTec AD, 3 of 28 VisA, 11 of 28 UCF-Crime, and 19 of 28 XD-Violence \DEC configurations.
No \PROB configuration exceeds it.
An exceedance requires a difference greater than $10^{-12}$ on the unit-interval scale.
These measurements support the fixed-score comparison in \Cref{sec:ties}.

\end{document}